\documentclass[10pt,journal,compsoc]{IEEEtran}
\ifCLASSOPTIONcompsoc
  \usepackage[nocompress]{cite}
\else
  \usepackage{cite}
\fi

\ifCLASSINFOpdf
\else
\fi

\usepackage{graphicx}
\usepackage{amsmath}
\usepackage{amssymb}
\usepackage{booktabs}
\usepackage{tabulary}
\usepackage{multirow}
\usepackage{overpic}
\usepackage{xspace}
\usepackage{epsfig}
\usepackage{hyperref}
\usepackage{graphicx}
\usepackage{bbm}
\usepackage[ruled]{algorithm2e}
\usepackage{algorithmic}
\usepackage{colortbl}
\graphicspath{ {./figures/} }
\usepackage{epsfig}
\usepackage{hyperref}
\usepackage{graphicx}
\usepackage{amsmath}
\usepackage{amssymb}
\usepackage{booktabs}
\usepackage{tabulary}
\usepackage{multirow}
\usepackage{overpic}
\usepackage{xspace}
\usepackage[ruled]{algorithm2e}
\usepackage{algorithmic}
\usepackage{bbm}
\usepackage[flushleft]{threeparttable}
\newcommand{\vct}[1]{\boldsymbol{#1}} 
\newcommand{\mat}[1]{\boldsymbol{#1}} 

\newcommand{\methodname}{\text{ProPos}\xspace}

\newcommand{\app}{\raise.17ex\hbox{$\scriptstyle\sim$}}

\newcommand{\ie}{\textit{i}.\textit{e}.\xspace}
\newcommand{\eg}{\textit{e}.\textit{g}.\xspace}
\newcommand{\std}[1]{{{\scriptsize $\pm$#1}}} 
\newcommand{\tabincell}[2]{\begin{tabular}{@{}#1@{}}#2\end{tabular}}

\newlength\savewidth\newcommand\shline{\noalign{\global\savewidth\arrayrulewidth
  \global\arrayrulewidth 1pt}\hline\noalign{\global\arrayrulewidth\savewidth}}
\newcommand{\lossname}{{PSL}\xspace}
\newcommand{\T}{^{\textrm T}} 
\DeclareMathOperator*{\argmax}{arg\,max}
\DeclareMathOperator*{\argmin}{arg\,min}

\definecolor{DarkGreen}{rgb}{0.1,0.5,0.1}
\definecolor{DarkRed}{rgb}{0.5,0.1,0.1}
\definecolor{DarkBlue}{rgb}{0.1,0.1,0.5}
\definecolor{Gray}{rgb}{0.2,0.2,0.2}

\hypersetup{
          breaklinks=true,   
          colorlinks=true,   
          pdfusetitle=true,  
      }

\begin{document}

\title{Learning Representation for Clustering via Prototype Scattering and Positive Sampling}
\author{Zhizhong~Huang,~Jie~Chen,~Junping~Zhang,~
        and~Hongming~Shan
\IEEEcompsocitemizethanks{\IEEEcompsocthanksitem Z. Huang, J. Chen, and J. Zhang are with the Shanghai Key Lab of Intelligent Information Processing and the School of Computer Science, Fudan University, Shanghai 200433, China.\protect\\ 
Email: \{zzhuang19, chenj19, jpzhang\}@fudan.edu.cn.
\IEEEcompsocthanksitem H. Shan is with  Institute of Science and Technology for Brain-inspired Intelligence,  MOE Frontiers Center for Brain Science, and Key Laboratory of Computational Neuroscience and Brain-Inspired Intelligence, Fudan University, Shanghai, 200433, China,  and also with  Shanghai Center for Brain Science and Brain-inspired Technology, Shanghai 201210, China.\protect\\ 
E-mail: hmshan@fudan.edu.cn.}
\thanks{Manuscript received xx xx, 2022; revised xx xx, 2022.}}

\markboth{IEEE Transactions on Pattern Analysis and Machine Intelligence,~Vol.~xx, No.~xx, xx~2022}%
{Huang \MakeLowercase{\textit{et al.}}: Bare Demo of IEEEtran.cls for Computer Society Journals}

\IEEEtitleabstractindextext{%
\begin{abstract}
Existing deep clustering methods rely on either contrastive or non-contrastive representation learning for downstream clustering task. Contrastive-based methods thanks to negative pairs learn uniform representations for clustering, in which negative pairs, however, may inevitably lead to the class collision issue and consequently compromise the clustering performance. Non-contrastive-based methods, on the other hand, avoid class collision issue, but the resulting non-uniform representations may cause the collapse of clustering. To enjoy the strengths of both worlds, this paper presents a novel end-to-end deep clustering method with prototype scattering and positive sampling, termed \methodname. Specifically, we first maximize the distance between prototypical representations, named prototype scattering loss, which improves the uniformity of representations. Second, we align one augmented view of instance with the sampled neighbors of another view---assumed to be truly positive pair in the embedding space---to improve the within-cluster compactness, termed positive sampling alignment. The strengths of \methodname are avoidable class collision issue, uniform representations, well-separated clusters, and within-cluster compactness. By optimizing \methodname in an end-to-end expectation-maximization framework, extensive experimental results demonstrate that \methodname achieves competing performance on moderate-scale clustering benchmark datasets and establishes new state-of-the-art performance on large-scale datasets. Source code is available at \url{https://github.com/Hzzone/ProPos}.
\end{abstract}

}

\maketitle

\IEEEdisplaynontitleabstractindextext

\IEEEpeerreviewmaketitle

\bstctlcite{IEEEexample:BSTcontrol}

\section{Introduction}

Deep clustering is gaining considerable attention as it aims to learn the representation of images and perform clustering in an end-to-end fashion. The main thrust to advance deep clustering is the self-supervised representation learning, including contrastive learning~\cite{he2020momentum,chen2020simple} and non-contrastive learning~\cite{grill2020bootstrap,chen2021exploring}. 

Remarkably, existing deep clustering methods heavily rely on contrastive representation learning, referred to as contrastive-based methods~\cite{wang2021unsupervised,van2020scan,li2020prototypical,li2021contrastive,tao2021clustering,tsai2020mice,niu2021spice}.
Specifically, they are usually built upon MoCo~\cite{he2020momentum} or SimCLR~\cite{chen2020simple}, requiring specially designed losses~\cite{wang2021unsupervised,li2020prototypical,li2021contrastive,tao2021clustering,tsai2020mice} or an extra pre-training stage for more discriminative representations~\cite{van2020scan,niu2021spice}.
Although achieving promising clustering results, contrastive-based methods usually require a large number of negative examples to learn uniform representations in an embedding space where all instances are well-separated.
The involved negative pairs may \emph{inevitably} lead to the class collision issue that different instances from the same semantic class are regarded as negative pairs and are \emph{wrongly} pushed away, which hampers the downstream clustering. An alternative perspective on this issue is to separate the typical contrastive loss into two terms~\cite{wang2020understanding}:
1) \emph{alignment} term to improve the closeness of positive pairs, and 
2) \emph{uniformity} term to encourage instances to be uniformly distributed on a unit hypersphere by pushing away the negative pairs.
Apparently, the uniformity term could introduce class collision issue~\cite{arora2019theoretical} as the constructed negative pairs may not be \emph{truly} negative.

Different from contrastive learning,  non-contrastive learning \emph{only} involves the alignment term using the representations of one augmented view to predict another. The non-contrastive learning can avoid the class collision issue as there are no negative pairs. 
Lacking the uniformity term in contrastive loss, it is not guaranteed to learn uniform representations~\cite{wang2020understanding,zhang2022does}, which may cause the collapse of downstream clustering---most samples are assigned to few clusters.
This phenomenon would even worsen when learning in conjunction with extra clustering losses introduced by current state-of-the-art deep clustering methods~\cite{li2020prototypical,li2021contrastive}; see Sec.~\ref{sec:ablation_study} and supplemental Fig.~\ref{fig:pcl_moco_and_byol}.

To enjoy the strengths of both worlds, we propose a novel end-to-end deep clustering method, \methodname, with two novel techniques: prototype scattering loss and positive sampling alignment.
First, considering that different prototypes/clusters are \emph{truly} negative pairs, we propose to perform contrastive learning over prototypical representations, in which two augmented views of the same prototypes are positive pairs and different prototypes are negative pairs. This yields the proposed prototype scattering loss or \lossname, which maximizes the between-cluster distance so as to learn uniform representations towards well-separated clusters.
Second, to improve the within-cluster compactness, we further propose to align one augmented view of the instance with the randomly sampled neighbors of another view that are assumed to be \emph{truly} positive pairs in the embedding space, which we refer to as positive sampling alignment or PSA. Compared to conventional alignment between two augmented views, the proposed PSA takes into account the neighboring samples in the embedding space, improving the within-cluster compactness. 
Moreover, we optimize \methodname in an expectation-maximization~(EM) framework, in which we iteratively perform E-step as estimating the instance pseudo-labels via spherical $k$-means and M-step as minimizing the proposed losses. 

The contributions are summarized as follows:
\begin{itemize}

\item We propose a novel end-to-end deep clustering method, termed \methodname, which enjoys the advantages of contrastive and non-contrastive representation learning: avoidable class collision issue, uniform representations for improved clustering stability, well-separated clusters, and improved within-cluster compactness.
\item We propose a novel prototype scattering loss or \lossname, which can align one augmented view of prototypes with another view and maximize the between-cluster scattering on the unit hypersphere, hence maximizing the inter-cluster distance for uniform representations.
\item We propose a positive sampling alignment or PSA to extend instance alignment by taking into account neighboring positive examples in the embedding space, which can improve the within-cluster compactness.

\item By optimizing \methodname in an EM framework, extensive experimental results on several benchmark datasets demonstrate that \methodname outperforms the existing state-of-the-art methods by a significant margin, especially for large-scale datasets.
\end{itemize}

The remainder of this paper is organized as follows. A brief review on the related work of self-supervised learning and deep clustering is given in Sec.~\ref{sec:related}, followed by the contrastive and non-contrastive representation learning in Sec.~\ref{sec:preliminary}. We present the proposed PSL and PSA as well as our \methodname in Sec.~\ref{sec:method}. Experimental results are reported and analyzed in Sec.~\ref{sec:exp}, where Sec.~\ref{sec:justification} further justifies the motivation. Finally, Sec.~\ref{sec:discuss} discusses the relations to previous works, followed by  a concluding summary in Sec.~\ref{sec:conc}.
\section{Related Work}\label{sec:related}

This section briefly surveys the development of self-supervised learning and deep clustering.

\subsection{Self-Supervised Learning}
Previous self-supervised learning~(SSL) methods for representation learning attempt to capture the data distribution using generative models~\cite{donahue2016adversarial,donahue2019large} or learn the representations through some special designed pretext tasks~\cite{doersch2015unsupervised,noroozi2016unsupervised,zhang2016colorful,caron2018deep}. 
In recent years, contrastive learning methods~\cite{wu2018unsupervised,he2020momentum,chen2020simple} have shown promising results for both representation learning and downstream tasks.
Contrastive representation learning requires a large number of negative examples to achieve instance-wise discrimination in an embedding space where all instances are well-separated.
The constructed negative pairs usually require a large batch size~\cite{chen2020simple}, memory queue~\cite{he2020momentum}, or memory bank~\cite{wu2018unsupervised}, which not only bring extra computational cost but also give rise to class collision issue~\cite{saunshi2019theoretical} that the semantic similar instances are pushed away since they could be regarded as negative pairs.
For example, MoCo~\cite{he2020momentum} uses a memory queue to store the consistent representations output by a moving-averaged encoder.
However, the class collision issue remains unavoidable. Some attempts have been made to address this issue~\cite{khosla2020supervised,hu2021adco,chuang2020debiased}. 

On the contrary, the recent studies of SSL demonstrate that the negative examples are not necessary, termed non-contrastive methods~\cite{caron2020unsupervised,grill2020bootstrap,chen2020simple}. Recently, mask image modeling~(MIM) such as MAE~\cite{he2021masked} arises a new trend for self-supervised learning that leverages ViT~\cite{dosovitskiy2020image} to directly reconstruct mask images. However, MIM may not be ready for deep clustering yet as ViT needs to be trained on large datasets such as ImageNet-1k~\cite{deng2009imagenet} and it does not learn discriminative representations for deep clustering. 

In summary, SSL methods mainly focus on inducing transferable representations for the (supervised) downstream tasks instead of grouping the data into different semantic classes for deep clustering.

\subsection{Deep Clustering}
Deep clustering~\cite{xie2016unsupervised,chang2017deep,chang2018deep,haeusser2018associative,chang2019local,wu2019deep,ji2019invariant,han2021autonovel,li2022twin} has been significantly advanced by self-supervised representation learning. 
Most of deep clustering methods are based on contrastive learning by exploiting the discriminative representations, learned from contrastive learning, to assist the downstream clustering tasks~\cite{van2020scan,niu2021spice,li2022twin} or simultaneously optimize representation learning and clustering~\cite{tao2021clustering,tsai2020mice,li2020prototypical,shen2021you,guo2022hcsc}. For example, SCAN~\cite{van2020scan} yields the confident pseudo-labels by the pre-trained SimCLR model, and IDFD~\cite{tao2021clustering} proposes to perform both instance discrimination and feature decorrelation. 
Although GCC~\cite{zhong2021graph} and WCL~\cite{zheng2021weakly} select the neighbor samples from a graph as pseudo-positive examples for contrastive loss, however, they still suffer from the class collision issue as these selected examples may not be \emph{truly} positive. In a nutshell, all of them are built upon the contrastive learning framework, in which they require a large number of negative examples to maintain uniform representations, inevitably leading to class collision issue.

Although some attempts have been made to use non-contrastive learning, such as BYOL~\cite{grill2020bootstrap}, to avoid class collision issue~\cite{lee2020learning,regatti2021consensus}, they produce inferior results as they suffer from the collapse of downstream clustering due to the non-uniform representations.

Our work advances deep clustering via the two novel techniques to address current drawbacks. First, the proposed \lossname maximizes the between-cluster distance, which leads to uniform representations, hence alleviating the collapse of downstream clustering. Second, the proposed positive sampling alignment improves within-cluster compactness. As a result, the proposed \methodname can enjoy the strengths of both worlds: avoidable class collision issue, uniform representation for improved clustering stability, well-separated clusters, and improved within-cluster compactness. 
In Sec.~\ref{sec:discuss}, we discuss the differences from existing methods including CC~\cite{li2021contrastive}, GCC~\cite{zhong2021graph}, WCL~\cite{zheng2021weakly}, PCL~\cite{li2020prototypical}, and instance-reweighted contrastive loss~\cite{mitrovic2020representation}.
\section{Preliminary}
\label{sec:preliminary}
Here, we briefly introduce representative contrastive learning and non-contrastive learning methods. 

\subsection{Contrastive Learning} 
Contrastive learning methods perform instance-wise discrimination~\cite{wu2018unsupervised} using the InfoNCE loss~\cite{oord2018representation}. Formally, assume that we have one instance $\vct{x}$, its augmented version $\vct{x}^+$ by using random data augmentation, and a set of $M$ negative examples drawn from the dataset, $\{\vct{x}_1^{-},\vct{x}_2^{-},\ldots,\vct{x}_M^{-}\}$. The contrastive learning aims to  learn an embedding function $f(\cdot)$ that maps $\mat{x}$ onto a unit hypersphere. The corresponding InfoNCE loss for one instance is defined as follows:
\begin{align}
 &  -\log \frac{\exp\left( \frac{f(\mat{x})\T f(\mat{x}^{+})}{\tau}\right)}{\exp\left(\frac{f(\mat{x})\T f(\mat{x}^{+})}{\tau}\right)\!+\!\sum\limits_{i=1}^{M} \exp\left(\frac{f(\mat{x})\T f(\mat{x}_{i}^{-})}{\tau}\right)} \label{eq:contr_loss}\\ 
 \approx& \underbrace{- \frac{f(\mat{x})\T f(\mat{x}^{+})}{\tau}}_{\text{instance alignment}}
         + \underbrace{\log \sum_{i=1}^{M} \exp\left(\frac{f(\mat{x})\T f(\mat{x}_{i}^{-})}{\tau}\right)}_{\text{instance uniformity}},\label{eq:contr_decouple_loss}
\end{align}
where the representation $f(\vct{x})$ is $\ell_2$ normalized on a unit hypersphere, and the temperature $\tau$ controls the concentration level of representations.

Intuitively, the InfoNCE loss aims to pull together the positive pair ($\mat{x}, \mat{x}^+$) from two different data augmentations of the same instance, and push $\mat{x}$ away from $M$ negative examples of other instances. As discussed in~\cite{wang2020understanding}, the InfoNCE loss in Eq.~\eqref{eq:contr_loss} can be approximately decoupled into two terms in Eq.~\eqref{eq:contr_decouple_loss}: the first term refers to as \emph{instance alignment}, and the second term \emph{instance uniformity}. Despite the alignment term pulls together the positive pair, the key to avoiding representation collapse is the uniformity term, which makes the negative examples uniformly distributed on the unit hypersphere. Although alleviating the collapse of downstream clustering, the negative examples may inevitably lead to the class collision issue~\cite{arora2019theoretical}, hurting the representations for clustering.

\subsection{Non-Contrastive Learning}
Non-contrastive methods only optimize the alignment term in Eq.~(\ref{eq:contr_decouple_loss}) to match the representations between two augmented views~\cite{grill2020bootstrap,chen2021exploring}. Specifically, they often leverage an online, a target, and a predictor network to bridge the gap between these two views with stop gradient operation to avoid representation collapse. In particular, if $\tau=0.5$, the loss used in~\cite{grill2020bootstrap,chen2021exploring} can be written as:
\begin{align}
     - 2g\left(f(\mat{x})\right)\T f^{\prime}(\mat{x}^{+}) \!=\! \left\|g\left(f(\mat{x})\right) \!-\! f^{\prime}(\mat{x}^{+})\right\|_{2}^{2} - \text{2},
    \label{eq:non_contr_loss}
\end{align}
where $g(\cdot)$, $f(\cdot)$, and $f'(\cdot)$ are  the predictor, online, and target networks, respectively; $g(f(\vct{x}))$ and $f^{\prime}(\vct{x}^+)$ are $\ell_2$-normalized. 
Without using negative pairs, non-contrastive learning avoids the class collision issue.
However, due to the lack of uniformity, they tend to produce non-uniform representations that usually result in the collapse of downstream clustering, making them unstable for deep clustering; see Sec.~\ref{sec:justification}.
\section{Method}
\label{sec:method}
The goal of deep clustering is to learn the representation of images and perform the clustering task simultaneously. Our \methodname advances 
deep clustering via two novel techniques: prototype scattering loss (PSL) and positive sampling alignment (PSA) detailed in Secs.~\ref{sec:proto_cl} and~\ref{sec:positive_sampling}, respectively. 
We then present the overview of \methodname and its EM optimization process in Sec.~\ref{sec:method_em}.

\begin{figure}[t]
    \centering
    \includegraphics[width=1.0\linewidth]{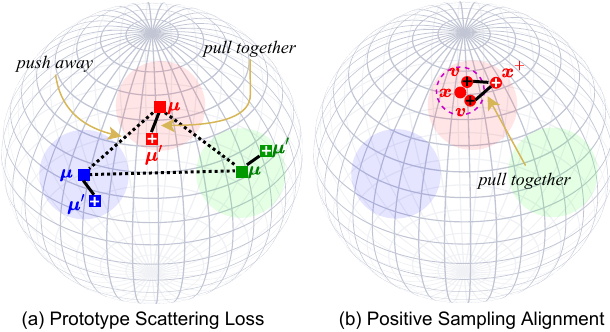}
    \caption{Illustration of the proposed two key techniques in \methodname. (a) The proposed prototype scattering loss to encourage alignment and maximize the between-cluster distance. (b) The proposed positive sampling alignment to encourage the alignment between sampled neighbors of one view with another for better within-cluster compactness.\label{fig:illustration}
    }
\end{figure}

\subsection{Prototype Scattering Loss}\label{sec:proto_cl}

A good clustering is supposed to have well-separated prototypes/clusters. Assuming that the dataset has $K$ clusters, where $K$ is a predefined number and assumed to be known, it naturally constructs a contrastive loss for these $K$ prototypes as for one prototype, the remaining $K-1$ prototypes are definitely negative examples. Therefore, we propose a prototype scattering loss or \lossname, which encourages the prototypical alignment between two augmented views and the prototypical uniformity, hence maximizing the inter-cluster distance.

Specifically, assume we obtain $K$ prototypes from one view in the embedding space, $\{\vct{\mu}_1, \vct{\mu}_2, \ldots, \vct{\mu}_K\}$, and another $K$ prototypes from another view, $\{\vct{\mu}_1^\prime, \vct{\mu}_2^\prime, \ldots, \vct{\mu}_K^\prime\}$, our proposed \lossname, illustrated in Fig.~\ref{fig:illustration}(a), is defined as follows: 
\begin{align}
\mathcal{L}_{\mathrm{\MakeLowercase{\lossname}}} &= \frac{1}{K}\sum_{k=1}^{K} -\log \frac{\exp\left(\frac{\vct{\mu}_k\T \vct{\mu}_k^\prime}{ \tau}\right)}{\exp\left(\frac{\vct{\mu}_k\T \vct{\mu}^\prime_k}{\tau}\right) \!+\! \sum\limits_{\substack{j=1\\ j\neq k}}^{K}\exp\left(\frac{\vct{\mu}_k\T \vct{\mu}_j}{\tau}\right)}
    \label{eq:proto_contr} \\
&\approx \underbrace{\frac{1}{K}\sum_{k=1}^{K}- \frac{\vct{\mu}_k\T \vct{\mu}_k^\prime}{\tau}}_{\text{prototypical alignment}}+ \underbrace{\frac{1}{K}\sum_{k=1}^{K}\log \sum_{\substack{j=1\\ j\neq k}}^{K} \exp\left(\frac{\vct{\mu}_k\T \vct{\mu}_j}{\tau}\right)}_{\text{prototypical uniformity}}.
    \label{eq:decoupled_proto_contr}
\end{align}
Here, the cluster centers $\vct{\mu}_k$ and $\vct{\mu}_k^\prime$ are computed within a mini-batch $\mathcal{B}$ as follows:
\begin{align}
\vct{\mu}_{k} &=\frac{\sum_{\mat{x} \in \mathcal{B} }  p(k\vert \mat{x}) f(\mat{x})}{\|\sum_{\mat{x} \in \mathcal{B} }  p(k\vert \mat{x}) f(\mat{x})\|_2}, \label{eq:compute_centers_1} \\
\vct{\mu}_{k}^\prime &=\frac{\sum_{\mat{x} \in \mathcal{B}}  p(k\vert\mat{x}) f^\prime(\mat{x})}{\|\sum_{\mat{x} \in \mathcal{B}}  p(k\vert\mat{x}) f^\prime(\mat{x})\|_2},
\label{eq:compute_centers_2}
\end{align}
where $p(k\vert\mat{x})$ is the cluster assignment posterior probability.
When $K > |\mathcal{B}|$, it is obvious that the mini-batch cannot cover all clusters. To this end, we zero out the losses and logits of empty clusters for each iteration.
During training, it is critical to estimate accurate $p(k\vert\mat{x})$ to optimize the proposed \lossname, so we adopt an EM framework that alternately uses a $k$-means clustering for every epoch at the E-step, and then minimize Eq.~\eqref{eq:proto_contr} at the M-step, which will be detailed later.

Intuitively, our \lossname for prototypes is similar to conventional contrastive loss in Eq.~\eqref{eq:contr_loss} for instances. The key difference is that \lossname will not cause the class collision issue as the prototypes are definitely negative examples for each other, which is more suitable for deep clustering.
However, the cluster centers may not be as accurate as expected during the early epochs of training. For an accurate initialization, following~\cite{li2020prototypical}, \lossname will be involved in training after the finish of the learning rate warmup.

Similarly, \lossname in Eq.~(\ref{eq:proto_contr}) can be approximately divided into Eq.~(\ref{eq:decoupled_proto_contr}): \textit{prototypical alignment}  and \textit{prototypical uniformity}.
On one hand, the prototypical alignment is to align the prototypes between two views, which can stabilize the update of the prototypes. On the other hand, the prototypical uniformity is to encourage the prototypes to be uniformly distributed on a unit hypersphere, which can maximize the inter-cluster distance. We note that there are two cluster-level losses related to ours: ProtoNCE~\cite{li2020prototypical} to improve cluster compactness and CC~\cite{li2021contrastive} to contrast cluster assignments; we discuss the major differences in Sec.~\ref{sec:discuss}.

\noindent\textbf{Uniform representation and well-separated clusters.} \quad In the context of non-contrastive learning for deep clustering, the lack of uniformity term in Eq.~(\ref{eq:non_contr_loss}) fails to produce uniform representations, which may cause severe collapse of downstream clustering. \lossname overcomes this drawback by maximizing the inter-clusters distance between prototypical representations, which yields uniform representations and well-separated clusters.

\subsection{Positive Sampling Alignment}\label{sec:positive_sampling}

Following previous discussion, the negative examples are essential for contrastive-based deep clustering to learn uniform representations, at the cost of inevitable class collision issue~\cite{saunshi2019theoretical} that harms the within-cluster compactness. On the other hand, non-contrastive learning can avoid the class collision issue by \textit{only} optimizing the instance alignment. However, the conventional instance alignment in Eq.~\eqref{eq:non_contr_loss} only encourages the representation of one augmented view to be close to another view. In the context of clustering, such compactness is instance-wise and neutral for deep clustering, since the semantic class information cannot be captured at only instance level.

To improve within-cluster compactness for conventional instance alignment while avoiding class collision issue, we propose to optimize the \emph{opposite} of the uniformity instead, \ie the compactness within clusters. We aim to encourage the neighboring examples around one augmented view---sampled from the embedding space and assumed to be \emph{truly} positive pairs---to be aligned with another view, as shown in Fig.~\ref{fig:illustration}(b). Our motivation is that although we cannot guarantee the negative pairs constructed from the dataset are \emph{truly negative}, we can certainly assume that the neighboring samples around one view in the embedding space are \emph{truly positive} with respect to another view and belong to the same class. 
Therefore, we propose a positive sampling alignment to extend the instance alignment in Eq.~\eqref{eq:non_contr_loss} by taking into account the neighboring samples towards improved within-cluster compactness. 

The key step of PSA is to sample the neighboring examples $\vct{v}$. A natural way is modeling the representation of one augmented view of an instance as a continuous distribution in the embedding space. We thus introduce a Gaussian distribution thanks to its simplicity, which can be formulated as follows:
\begin{align}
\vct{v}\sim\mathcal{N}\left(f(\vct{x}), \sigma^2 \mat{I}\right),
    \label{eq:aug_distribution}
\end{align}
where $\mat{I}$ represents the identity matrix and $\sigma$ is a positive hyperparameter controlling how many samples around one view can be treated as positive pairs with another view. However, the sampled examples from Eq.~\eqref{eq:aug_distribution} cannot allow the error to be backpropagated through the network to update the network parameters. By leveraging the reparametrization trick~\cite{kingma2013auto}, the positive sampling can be implemented as follows:
\begin{align}
\vct{v} = f(\vct{x}) + \sigma \vct{\epsilon},\quad\text{where}\quad \vct{\epsilon}\sim\mathcal{N}\left(0, \mat{I}\right).
\end{align}
We then extend the instance alignment in Eq.~(\ref{eq:non_contr_loss}) by taking into account the neighboring samples. With only sampling one example from the Gaussian distribution, the positive sampling alignment (PSA) can be formally defined as:
\begin{align}
    \mathcal{L}_{\mathrm{psa}} &= \left\|g(\mat{v}) - f^{\prime}(\mat{x}^{+})\right\|_{2}^{2} \notag\\
    &= \left\|g\left(f(\mat{x}) + \sigma\vct{\epsilon}\right) - f^{\prime}(\mat{x}^{+})\right\|_{2}^{2}.
    \label{eq:aug_non_contr}
\end{align}
Here, when $\sigma=0$, PSA reduces to Eq.~\eqref{eq:non_contr_loss}.

\noindent\textbf{Avoidable class collision issue.}\quad 
In the context of non-contrastive learning without negative examples,
our PSA can guarantee that the positive examples around one instance, sampled in the embedding space, are from the same cluster and form \emph{truly positive} pairs. Therefore, optimizing PSA loss would not cause class collision issue that exists in contrastive-based deep clustering methods. We discuss the difference from~\cite{zhong2021graph,zheng2021weakly} that sample positive examples from the dataset in Sec.~\ref{sec:discuss}.

\noindent\textbf{Improved within-cluster compactness.}\quad Unlike conventional instance alignment for non-contrastive learning, our PSA in Eq.~(\ref{eq:aug_non_contr}) encourages neighboring examples around one augmented view---either different augmented examples of the same instance or same/different augmented examples of different instances within the same cluster---to be positive pairs with another view. This helps to improve within-cluster compactness.

\begin{figure*}[t]
    \centering
    \includegraphics[width=1\linewidth]{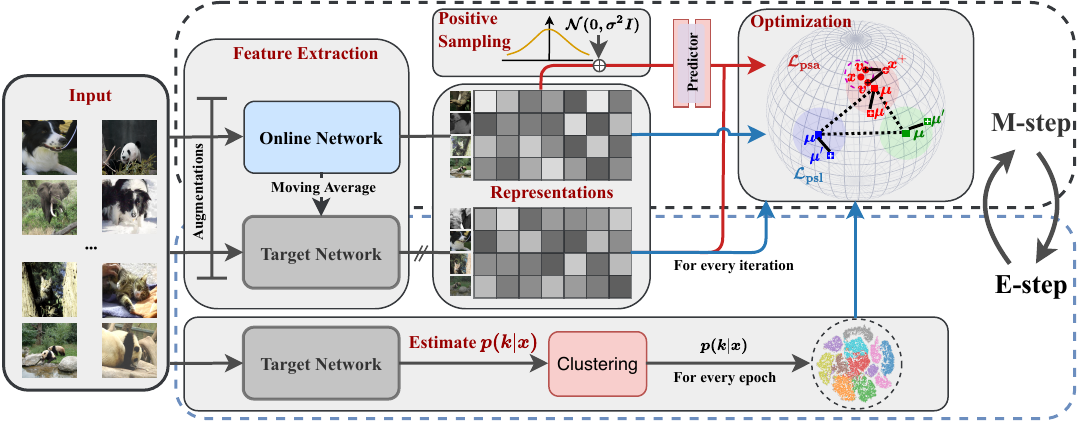}
    \caption{The overall framework of the proposed \methodname in an EM framework.}
    \label{fig:em_framework}
\end{figure*}

\subsection{Overview of \methodname and its optimization}
\label{sec:method_em}

We present the overview of our \methodname in Fig.~\ref{fig:em_framework} and optimize it in an EM framework to facilitate the understanding of the training procedure.

\subsubsection{Overview}
We build \methodname upon non-contrastive learning framework similar to BYOL, which is comprised of three networks: an online, a target, and a predictor. During training, the parameters of target network are momentum updated~(\textit{a.k.a} moving averaged) from the ones of online network, following
\begin{align}
\mat{\theta}_{\mathrm{target}} = m \mat{\theta}_{\mathrm{target}}+(1-m) \mat{\theta}_{\mathrm{online}},
\end{align}
where $m\in [0, 1)$ is the coefficient, and $\mat{\theta}_*$ denotes the parameters. Two different random data augmentations from the same inputs are fed into the network to optimize the proposed losses in an EM framework.
Specifically, $k$-means clustering is performed in the E-step at the beginning of each epoch to obtain the $p(k\vert \mat{x})$, which is fixed in the latter training to optimize the proposed \lossname.
The $\vct{\mu}_k$ and $\vct{\mu}_k^\prime$ are computed from online and target networks using Eqs.~\eqref{eq:compute_centers_1} and \eqref{eq:compute_centers_2}, respectively.
Furthermore, PSA is applied to the representations from online network, which are then passed through predictor network for alignment.

\subsubsection{EM framework}

The optimization of \methodname is done in an EM framework, where E-step and M-step are detailed as follows; supplemental Sec.~\ref{sec:em_framework} presents detailed derivations.

\noindent\textbf{E-step:}  
This step aims to estimate $p(k\vert \mat{x})$ for the proposed \lossname. We perform spherical $k$-means algorithm on the features extracted from the target network since the target network performs more stable and yields more consistent clusters, similar to BYOL and MoCo. Although we need an additional $k$-means clustering to obtain the cluster pseudo-labels $p(k\vert \mat{x})$ for every $r$ epochs, we found that even with a larger $r > 1$, our method can still produce consistent performance improvement over the baseline methods. Therefore, our method will not introduce much computation cost and is robust to the cluster pseudo-labels; see detailed results in Secs.~\ref{sec:hyperparameter_analysis} and \ref{sec:computational_cost}. 
Finally, with $p(k\vert \mat{x})$, we build the prototypical representations within a mini-batch without additional memory.\\
\noindent\textbf{M-step:}
Combining \lossname in Eq.~(\ref{eq:proto_contr}) and the PSA in Eq.~(\ref{eq:aug_non_contr}) yields our objective function for M-step as follows:
\begin{equation}
    \mathcal{L} = \mathcal{L}_{\mathrm{psa}} + \lambda_{\mathrm{\MakeLowercase{\lossname}}} \mathcal{L}_{\mathrm{\MakeLowercase{\lossname}}},
    \label{eq:overall_loss}
\end{equation}
where $\lambda_{\mathrm{\MakeLowercase{\lossname}}}$ controls the balance between two loss components. Therefore, there are only two hyper-parameters in the loss function, including: $\sigma$ in $\mathcal{L}_{\mathrm{psa}}$ and the loss weight $\lambda_{\mathrm{\MakeLowercase{\lossname}}}$; see the effects of hyper-parameters in Sec.~\ref{sec:hyperparameter_analysis}. 

\begin{algorithm}[t]\label{alg}
    \caption{Training Algorithm}
    \textbf{Input:}\hspace{0mm} Dataset $\mathcal{D}=\{\mat{x}\}$;\\ 
    \hspace{9.5mm} Functions $f(\cdot)$ and $f^\prime(\cdot)$ \\
    \textbf{Output:}\hspace{0mm} Clustering results $\{p(k|\mat{x})\}$.\\
    \Repeat{reaching max epochs}{
    E-step: update $\{p(k|\mat{x})\}$ for each sample in $\mathcal{D}$ using $k$-means clustering \\
    M-step:
    \Repeat{an epoch finished}{
        Randomly sample a mini-batch $\mathcal{B}$ from $\mathcal{D}$\\

        \For{each $\mat{x}_i$ in $\mathcal{B}$}{
            Randomly augment $\mat{x}$ and $\mat{x}^+$ \\
            Compute cluster centers using Eqs.~\eqref{eq:compute_centers_1} and ~\eqref{eq:compute_centers_2} \\
            $\mathcal{L}_{\mathrm{psl}}\leftarrow$Eq.~\eqref{eq:proto_contr} \\
            $\mathcal{L}_{\mathrm{psa}}\leftarrow$Eq.~\eqref{eq:aug_non_contr}\\
        }
        $\mathcal{L}\leftarrow$Eq.~\eqref{eq:overall_loss}\\
        Update $f^{\prime}$ with momentum moving average \\
        Update $f$ with SGD optimizer \\
    }
    }
\end{algorithm}

The training procedure of the proposed \methodname is presented in Algorithm~\ref{alg}.
 
\noindent\textbf{Relations between \lossname and PSA.}\quad 
A good clustering model should have well-separated clusters and within-cluster compactness. On the one hand, \lossname encourages well-separated clusters by maximizing inter-cluster distance, which, however, cannot improve within-cluster compactness. On the other hand, PSA can pull together the sampled neighboring examples around one augmented view and another view, which can further improve within-cluster compactness. By combing these two losses in Eq.~(\ref{eq:overall_loss}), we expect \methodname can improve deep clustering towards well-separated clusters and within-cluster compactness.
\section{Experiments}
\label{sec:exp}

\subsection{Experimental Setup}

\begin{table}[t]
  \centering
  \caption{Summary of the datasets.}
  \label{tab:dataset}
  \scalebox{0.95}
  {%
  \begin{tabular}{llrrr}
    \shline
    \textbf{Dataset} & \textbf{Split}      & \# \textbf{Samples} & \# \textbf{Classes} & \textbf{Image Size} \\
    \midrule
    CIFAR-10                    & Train+Test & 60,000 & 10   & 32$\times$32 \\
    CIFAR-20                    & Train+Test & 60,000  & 20  & 32$\times$32 \\
    STL-10                      & Train+Test & 13,000 & 10   & 96$\times$96 \\
    ImageNet-10                 & Train      & 13,000 & 10   & 96$\times$96  \\
    ImageNet-Dogs               & Train      & 19,500   & 15   & 96$\times$96  \\
    Tiny-ImageNet               & Train      & 100,000  & 200   & 224$\times$224  \\
    ImageNet-1k                    & Train      & 1,281,167                   & 1,000                        & 224$\times$224  \\
    \shline
    \end{tabular}
  }
\end{table}

\subsubsection{Datasets}
We conducted experiments on seven benchmark datasets, including \textbf{CIFAR-10}~\cite{krizhevsky2009learning}, \textbf{CIFAR-20}~\cite{krizhevsky2009learning}, \textbf{STL-10}~\cite{coates2011analysis}, \textbf{ImageNet-10}~\cite{chang2017deep}, \textbf{ImageNet-Dogs}~\cite{chang2017deep}, \textbf{Tiny-ImageNet}~\cite{le2015tiny}, and \textbf{ImageNet-1k}~\cite{deng2009imagenet}, which are summarized in Table~\ref{tab:dataset}. We note that CIFAR-20 contains 20 superclasses of CIFAR-100. STL-10 includes extra unlabeled images. ImageNet-10, ImageNet-Dogs, and Tiny-ImageNet are the widely-used subsets of ImageNet-1k~\cite{deng2009imagenet}, containing 10, 15, 200 classes, respectively. This paper follows the experimental settings widely used in deep clustering work~\cite{chang2017deep,wu2019deep,ji2019invariant,tsai2020mice,tao2021clustering}, including the image size, backbone, and train-test split. For image size, we use $32\times 32$ for CIFAR-10 and CIFAR-20, $96\times 96$ for STL-10, ImageNet-10, and ImageNet-Dogs, and $224\times 224$ for Tiny-ImageNet and ImageNet-1k. For train-test split, we use the whole datasets including training and testing set for CIFAR-10 and CIFAR-20 while both labeled and unlabeled data are employed for STL-10.

\subsubsection{Backbones}
We use ResNet-34~\cite{he2016deep} as the backbone for fair comparisons to report the main results on moderate-scale benchmark datasets. We use ResNet-18~\cite{he2016deep} on Tiny-ImageNet and ResNet-50~\cite{he2016deep} on ImageNet-1k, following the literature. Unless noted otherwise, we use ResNet-18 for the rest of the experiments. Since the image sizes of CIFAR-10 and CIFAR-100 are relatively small, following~\cite{chen2020simple}, we replace the first convolution layer of kernel size $7\times7$ and stride 2 with a convolution layer of kernel size $3\times 3$ and stride 1, and remove the first max-pooling layer for all experiments on CIFAR-10 and CIFAR-100.

\begin{figure*}[t]
  \centering
  \includegraphics[width=1.0\linewidth]{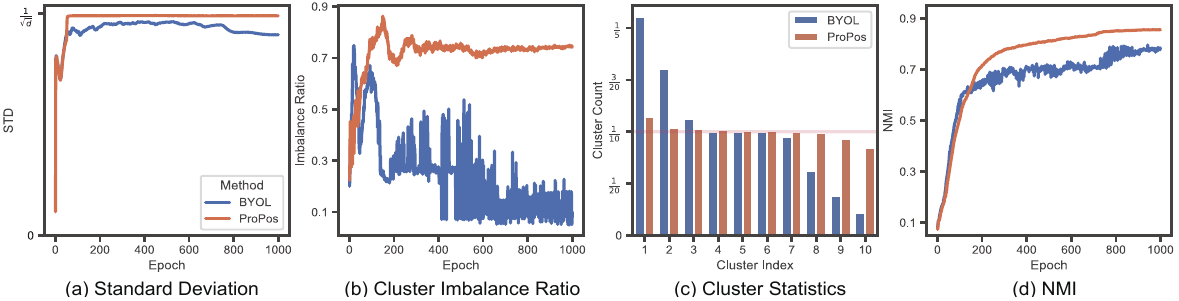}
  \caption{
  Detailed comparison between BYOL~\cite{grill2020bootstrap} and \methodname on CIFAR-10 in terms of (a) standard deviation~(STD) of $\ell_2$-normalized features to evaluate the uniformity, (b) cluster imbalance ratio computed by $\min(\{N_k\}_{k=1}^K)/\max(\{N_k\}_{k=1}^K)$ to show how balanced the clusters are,  (c) cluster statistics, or the sorted number of samples in each cluster for the model at 1000-th epoch, and (d) normalized mutual information~(NMI) between the clustering results and ground-truth labels.
  }
  \label{fig:training_vis_clusters}
\end{figure*}

\subsubsection{Implementation details}
\label{sec:implementation_details}
We train all methods with 1,000 epochs, strictly following the literature~\cite{tsai2020mice,tao2021clustering}, and adopt the stochastic gradient descent (SGD) optimizer and the cosine decay learning rate schedule with 50 epochs for learning rate warmup. The base learning rate for MoCo v2~\cite{chen2020improved}, BYOL~\cite{grill2020bootstrap}, and \methodname were 0.05, scaled linearly with the batch size (LearningRate = 0.05$\times$BatchSize/256). Note that the learning rates for predictor networks of BYOL and \methodname are $10\times$ as the learning rate of feature extractor. It is relatively important to achieve satisfactory performance, as discussed in~\cite{grill2020bootstrap,chen2021exploring}. 

For other hyperparameters of \methodname, the temperature~$\tau$, $\lambda_{\mathrm{psl}}$ for prototypical scattering loss, and $\sigma$ for positive sampling were set as $0.5$, $0.1$, and $0.001$, respectively. The mini-batch size was 512 for MoCo and 256 for the remaining methods. \methodname was trained on 4 NVIDIA V100 GPUs. 

Regarding CC~\cite{li2021contrastive} and PCL~\cite{li2020prototypical}, we tried our best to reproduce their results for fair comparisons. For CC, we used their official code. For PCL, under the fair conditions of MoCo, we set the loss weight of ProtoNCE to $0.01$ and the number of clusters to $\{250, 500, 1000\}$ following the suggestions of authors, which we found can achieve the best results. We integrated CC and PCL into BYOL by adding their losses without changing other settings.

\subsubsection{Configurations of SSL frameworks}
We adopt the same data augmentations as SimCLR~\cite{chen2020simple}, including ResizedCrop, ColorJitter, Grayscale, and HorizontalFlip. We have removed GaussianBlur since we only used a small image size for all datasets. We also strictly follow the settings of BYOL~\cite{grill2020bootstrap}. Specifically, despite the standard ResNet backbones, the projection and predictor networks have the architectures of FC-BN-ReLU-FC, where the projection dimension and hidden size were 256 and 4096 for both two networks, respectively.
For fair comparisons, we have also set the projection dimension of MoCo v2 as 256. We have used symmetric loss for all methods,
\ie, swapping two data augmentations to compute twice loss. For the momentum hyperparameter $m\in [0, 1)$, we set it to 0.996 for both BYOL and \methodname same as~\cite{grill2020bootstrap} and 0.99 for MoCo v2. For MoCo v2, the queue size, temperature for InfoNCE loss, weight decay were 4,096, 1.0, and $1.0\times10^{-4}$, respectively. We have not employed SyncBN in \methodname. We note that SyncBN would introduce much additional computation cost. Instead, we adopt the shufflingBN in MoCo to avoid the trivial solution of non-contrastive learning. 

\subsection{Empirical Justification}
\label{sec:justification}

We provide an empirical justification on how the proposed \methodname improves representation learning for deep clustering from the aspects of PSL and PSA.

\subsubsection{The role of PSL} 
The non-uniform representations produced by non-contrastive representation learning would lead to the collapse of downstream clustering where most samples are assigned to few clusters.
We emphasize that it is desirable to avoid the trivial solution for deep clustering. For example, CC~\cite{li2021contrastive} and etc~\cite{niu2020gatcluster,zhong2021graph} have usually employed an entropy term in loss function to regularize the model equally assigning the images into different clusters.
This paper mainly investigates this phenomenon of non-contrastive representation learning for deep clustering, and the theoretical analysis can be found in~\cite{zhang2022does,tian2021understanding}.

Our \methodname can encourage cluster uniformity for representations via \lossname.
Here, we use the representative non-contrastive learning method, Bootstrap Your Own Latent (BYOL), as the representation learning for deep clustering. We performed spherical $k$-means on the learned representations for the clustering task with 10 different initializations. Following~\cite{chen2021exploring}, we use the standard deviation (STD) of $\ell_2$-normalized representations to measure the uniformity. Ideally, if the $\ell_2$-normalized representations are uniformly distributed on a unit hypersphere, we have $\operatorname{STD}\left[\vct{z}^{\prime}\right] \approx 1 / \sqrt{d}$, where $\vct{z}^{\prime}$ and $d$ are the $\ell_2$-normalized version and the dimension of the feature representation $\vct{z}$. To justify the effectiveness of \methodname, we conducted the following experiments in terms of the uniformity of the representations, the collapse of clustering, and the clustering performance.

First, we visualize the uniformity of representations in Fig.~\ref{fig:training_vis_clusters}(a). Taking a look at STDs during the training stage, \methodname produces higher STDs, while BYOL performs unstable with the STDs gradually decreasing.
Note that a higher standard deviation close to $1 / \sqrt{d}$ indicates more uniform representations. Our \methodname yields more uniform representations than BYOL. Most importantly, the uniformity of \methodname is rather stable during training.  

Second, we further visualize the cluster imbalance to measure the potential collapse during clustering. More specifically, we compute the cluster imbalance ratio between the cluster with least samples and the cluster with most samples, $\min(\{N_k\}_{k=1}^K)/\max(\{N_k\}_{k=1}^K)$, where $N_k$ is the number of samples in $k$-th cluster.  A higher value indicates more balanced clusters. In addition, we also show the cluster statistics or the sorted number of samples in each cluster for the model at 1000-th epoch. The results of cluster imbalance during training and the cluster statistics at the final epoch are shown in Fig.~\ref{fig:training_vis_clusters}(b) and (c).   Fig.~\ref{fig:training_vis_clusters}(b) shows that the $k$-means clustering process of \methodname produces more balanced clusters with a higher cluster imbalance ratio. On the contrary, the clusters of BYOL are highly imbalanced, which is consistent with  decreasing STDs.
Moreover, Fig.~\ref{fig:training_vis_clusters}(c) shows that the cluster statistics for \methodname are approximately and equally assigned to different clusters, compared to almost long-tailed assignments of BYOL.
The more balanced clusters validate that \methodname can alleviate the collapse of k-means clustering over BYOL.

Finally, Fig.~\ref{fig:training_vis_clusters}(d) shows the clustering performance comparison between the proposed \methodname and BYOL, which is measured by the normalized mutual information (NMI) between clustering results and ground-truth labels. We can see that \methodname produces higher and more stable NMIs than BYOL. In line with the analysis above, we conclude that directly applying BYOL to deep clustering, although avoiding class collision issue, suffers from the collapse of $k$-means clustering due to the non-uniform representations. In contrast, \methodname with PSL yields more uniform representations and well-clustered samples.

\subsubsection{The role of PSA} 

\begin{figure}[t]
  \centering
  \includegraphics[width=1.0\linewidth]{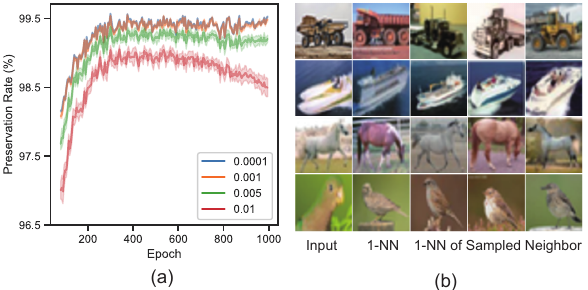}
  \caption{
    Visualization of sampled neighbors for \methodname on CIFAR-10: (a) The odds of sampled neighbors that have preserved their original semantic classes under different $\sigma$ during training; and (b) The input images, the 1-nearest-neighbors~(1-NN) of input images, and three sample neighbors at 100th epoch.
  }
  \label{fig:just_psa}
\end{figure}

\begin{table*}[t]
  \centering
  \caption{Clustering results (\%) of various methods on five benchmark datasets. The best and second best results are shown in bold and underline, respectively. We split different methods according to different training paradigms. The works most related to our method are IDFD and PCL that improve the representations for clustering.
  }
  \label{tab:results}
  \begin{threeparttable}
  \begin{tabular*}{1\linewidth}{@{\extracolsep{\fill}}l*{19}{c}}
    \shline
    \multirow{3}{*}{\textbf{Method}} & \multicolumn{3}{c}{\textbf{CIFAR-10}} & \multicolumn{3}{c}{\textbf{CIFAR-20}} &\multicolumn{3}{c}{\textbf{STL-10}} & \multicolumn{3}{c}{\textbf{ImageNet-10}} & \multicolumn{3}{c}{\textbf{ImageNet-Dogs}}\\
    \cmidrule{2-16}
     & NMI & ACC & ARI & NMI & ACC & ARI & NMI & ACC & ARI & NMI & ACC & ARI & NMI & ACC & ARI\\
    \midrule
    IIC~\cite{ji2019invariant} & 51.3 & 61.7 & 41.1    & -    & 25.7 & -        & 43.1 & 49.9 & 29.5  & -    & -    & -          & -    & -    & -     \\
    DCCM~\cite{wu2019deep} & 49.6 & 62.3 & 40.8 & 28.5 & 32.7 & 17.3 & 37.6 & 48.2 & 26.2 & 60.8 & 71.0 & 55.5 & 32.1 & 38.3 & 18.2    \\

    PICA~\cite{huang2020deep} & 56.1 & 64.5 & 46.7    & 29.6 & 32.2 & 15.9     & -    & -    & -     & 78.2 & 85.0 & 73.3      & 33.6 & 32.4 & 17.9         \\
    \midrule
    SCAN~\cite{van2020scan} &  79.7 & 88.3 & 77.2 & 48.6 & 50.7 & 33.3 & 69.8 & 80.9 & 64.6 & - & - & - & - & - & -\\
    NMM~\cite{Dang_2021_CVPR} & 74.8 & 84.3 & 70.9 & 48.4 & 47.7 & 31.6 & 69.4 & 80.8 & 65.0 & - & - & - & - & - & -\\
    
    \midrule
    
    CC~\cite{li2021contrastive}$^1$ &  70.5 & 79.0 & 63.7 & 43.1 & 42.9 & 26.6 & \underline{76.4} & 85.0 & 72.6 & 85.9 & 89.3 & 82.2 & 44.5 & 42.9 & 27.4\\
    MiCE~\cite{tsai2020mice} & 73.7 & 83.5 & 69.8 & 43.6 & 44.0 & 28.0 & 63.5 & 75.2 & 57.5 & - & - & - & 42.3 & 43.9 & 28.6\\
    GCC~\cite{zhong2021graph} & 76.4 & 85.6 & 72.8 & 47.2 & 47.2 & 30.5 & 68.4 & 78.8 & 63.1 & 84.2 & 90.1 & 82.2 & 49.0 & 52.6 & 36.2\\
    TCL~\cite{li2022twin}$^1$ & \underline{81.9} & 88.7 & 78.0 & 52.9 & 53.1 & 35.7 & \textbf{79.9} & \textbf{86.8} & \textbf{75.7} & 87.5 & 89.5 & 83.7 & 62.3 & 64.4 & 51.6 \\
    TCC~\cite{shen2021you} & 79.0 & \underline{90.6} & 73.3 & 47.9 & 49.1 & 31.2 & 73.2 & 81.4 & 68.9 & 84.8 & 89.7 & 82.5 & 55.4 & 59.5 & 41.7 \\

    \midrule

    MoCo~\cite{he2020momentum} & 66.9 & 77.6 & 60.8 & 39.0 & 39.7 & 24.2 & 61.5 & 72.8 & 52.4 & - & - & - & 34.7 & 33.8 & 19.7\\
    SimSiam~\cite{chen2020simple} & 78.6 & 85.6 & 73.6 & 52.2 & 48.5 & 32.7 & 65.9 & 71.6 & 57.2 & 83.1 & 92.1 & 83.3 & 58.3 & 67.4 & 50.1 \\
    BYOL~\cite{grill2020bootstrap} & 81.7 & {89.4} & \underline{79.0} & \underline{55.9} & \underline{56.9} & \underline{39.3} & 71.3  & {82.5}  & 65.7  & 86.6 & 93.9 & 87.2 & \underline{63.5} & \underline{69.4} & \underline{54.8}\\

    \midrule
    IDFD~\cite{tao2021clustering} & 71.1 & 81.5 & 66.3 & 42.6 & 42.5 & 26.4 & 64.3 & 75.6 & 57.5 & \textbf{89.8} & \underline{95.4} & \underline{90.1} & 54.6 & 59.1 & 41.3\\
    PCL~\cite{li2020prototypical} & 80.2 & 87.4 & 76.6 & 52.8 & 52.6 & 36.3 & 71.8 & 41.0 & 67.0 & 84.1 & 90.7 & 82.2 & 44.0 & 41.2 & 29.9 \\

    \methodname~(\textbf{ours}) & \textbf{88.6} & \textbf{94.3} & \textbf{88.4} & \textbf{60.6} & \textbf{61.4} & \textbf{45.1} & 75.8 & \underline{86.7} & \underline{73.7} & \underline{89.6} & \textbf{95.6} & \textbf{90.6} & \textbf{69.2} & \textbf{74.5} & \textbf{62.7} \\
  \shline
\end{tabular*}
  \begin{tablenotes}
  \item[1] CC and TCL use a large image size  of $224\times 224$ for all datasets.
  
\end{tablenotes}
\end{threeparttable}
 
\end{table*}

PSA assumes that the sampled neighbors are truly positive examples with respect to another view, \ie, they belong to the same semantic classes. To validate this assumption, we investigate the behavior of PSA during training by checking whether the semantic classes of input examples have been changed. 

First, we perform $k$-NN classification to predict the classes of both inputs and their sampled neighbors from testing set, and then use the proportion of sampled neighbors that have preserved original classes as the preservation rate. We run the experiments 10 times with different $\sigma$ for PSA. As shown in Fig.~\ref{fig:just_psa}(a), even at the early training stage, the sampled neighbors well preserve original classes for $\sigma<0.005$. The preservation rate drops as expected for $\sigma=0.01$ since PSA may sample the instances from another cluster for large $\sigma$. 

Second, we showcase some randomly-selected input images whose nearest neighbors have been changed during the sampling procedure, Specifically, Fig.~\ref{fig:just_psa}(b) visualizes the input images, their nearest neighbors, and the sampled neighbors. We note that we replaced the images of sampled neighbors with their nearest neighbors since it is too hard to reconstruct the image from embedding space. Obviously, the sampled neighbors belong to the same class as the input images even at the early stage of training (\ie 100th epoch).

In summary, the results suggest that the sampled neighbors could be truly positive examples both quantitatively and qualitatively so that PSA can improve the within-cluster compactness.

\subsection{Main Results}

In this section, we evaluate \methodname with previous state-of-the-art clustering methods on various benchmark datasets. We divide these methods into 5 types: i) methods without using contrastive learning; ii) multi-stage methods requiring step-by-step pretraining or finetuning; iii) methods directly outputting the cluster assignments; iv) methods learning general representations, \eg, MoCo; and v) methods improving representation learning for clustering. We strictly follow the experimental settings of previous works~\cite{tsai2020mice,tao2021clustering} for fair comparisons. We reproduced PCL~\cite{li2020prototypical}, SimSiam~\cite{chen2020simple}, and BYOL~\cite{grill2020bootstrap} under the same conditions, and directly use the their learned representations for $k$-means clustering.

In terms of qualitative results of clustering, we visualize the learned representations by t-SNE~\cite{van2008visualizing} for four different training epochs throughout the training process in supplemental Fig.~\ref{fig:clustering_quality} and the outlier points produced by the model at 1000-th epoch on CIFAR-10 in supplemental Fig.~\ref{fig:outlier_points_vis}.
For fair comparisons, supplemental Table~\ref{tab:fair_results_with_imagesize_and_split} presents the results excluding the testing set for CIFAR-10/20 and using a larger image size for ImageNet-10/Dogs.

\subsubsection{Results on moderate-scale datasets}
The comparisons on five moderate-scale datasets are reported in Table~\ref{tab:results}. The results of MoCo are referred from~\cite{tsai2020mice}. For fair  comparison, we excluded SPICE~\cite{niu2021spice} in Table~\ref{tab:results} since it requires multiple pre-training stages. 
\methodname achieves significant performance improvement on all benchmark datasets, demonstrating the superiority of \methodname for deep clustering to capture the semantic class information.

\begin{table}[t]
  \centering
  \caption{
      Clustering results (\%) on Tiny-ImageNet. We trained \methodname using ResNet-18. 
  }
  \label{tab:tiny_imagenet}
  \begin{tabular*}{1\linewidth}{@{\extracolsep{\fill}}lrrr}
      \shline
      \multirow{3}{*}{\textbf{Method}} & \multicolumn{3}{c}{\textbf{Tiny-ImageNet}} \\
      \cmidrule{2-4}
        & NMI       & ACC       & ARI       \\
      \midrule
      DCCM~\cite{wu2019deep}   & 22.4      & 10.8      & 3.8       \\
      PICA~\cite{huang2020deep}   & 27.7      & 9.8      & 4.0       \\
      CC~\cite{li2021contrastive}   & 34.0      & 14.0      & 7.1       \\
      GCC~\cite{zhong2021graph}     & 34.7      & 13.8      & 7.5       \\
      MoCo~\cite{he2020momentum} & 34.2 & 16.0 & 8.0 \\
      PCL~\cite{li2020prototypical} & 35.0 & 15.9 & 8.7 \\
      SimSiam~\cite{chen2020simple} & 35.1 & \underline{20.3} & 9.4 \\
      BYOL~\cite{grill2020bootstrap} & \underline{36.5} & 19.9 & \underline{10.0}\\
      \methodname~(\textbf{ours})     & \textbf{40.5}      & \textbf{25.6}      & \textbf{14.3}      \\
      \shline
  \end{tabular*}
\end{table}
\begin{table}[t]
  \centering
  \caption{Clustering results (\%) on ImageNet-1k using ResNet-50.}\label{tab:result_imagenet}
  \begin{tabular*}{1\linewidth}{@{\extracolsep{\fill}}lr}
  \shline
  \textbf{Method}      & \textbf{AMI}   \\
  \midrule
  DeepCluster~\cite{caron2018deep}  & 28.1 \\
  MoCo~\cite{he2020momentum}        & 28.5 \\
  PCL~\cite{li2020prototypical}     & \underline{41.0} \\
  \methodname~(\textbf{ours})     & \textbf{52.5} \\
  \shline
  \end{tabular*}
\end{table}

\begin{table}[t]
  \centering
  \caption{
    Linear evaluation on ImageNet-1k dataset. We report the Top-1 classification accuracy (\%) by training a linear classifier; the results are adopted from corresponding papers. The upper group uses more fair conditions, \eg, backbone and training epoch.
    }
  \label{tab:imagenet_linear_classifier}
  \begin{tabular*}{1\linewidth}{@{\extracolsep{\fill}}llrrr}
  \shline
  \multirow{3}{*}{\textbf{Method}} & \multirow{3}{*}{\textbf{Backbone}} & \multicolumn{2}{c}{\textbf{Pre-training}}   & \multirow{3}{*}{\textbf{ACC}}     \\
  \cmidrule{3-4}
    &    & Batch size& Epochs    &      \\
  \midrule
  Jigsaw~\cite{noroozi2016unsupervised}   & AlexNet   & 256  & -    & 34.6 \\
  Rotation~\cite{gidaris2018unsupervised} & AlexNet   & 128  & 100  & 38.7 \\
  DeepCluster~\cite{caron2018deep}             & AlexNet   & 256  & 500  & 41.0 \\
  InstDisc~\cite{wu2018unsupervised} & ResNet-50 & 256  & 200  & 54.0 \\
  LocalAgg~\cite{zhuang2019local} & ResNet-50 & 128  & 200  & 60.2 \\
  CMC~\cite{tian2020contrastive} & ResNet-50 & -    & 200  & 66.2 \\
  SimCLR~\cite{chen2020simple}   & ResNet-50 & 256  & 200  & 64.3 \\
  MoCo~\cite{he2020momentum} & ResNet-50 & 256  & 200  & 60.6 \\
  MoCo v2~\cite{chen2020improved}   & ResNet-50  & 256 & 200  & 67.5 \\
  PCL~\cite{li2020prototypical}    & ResNet-50   & 256   & 200  & 67.6 \\
  IFND~\cite{chen2021incremental}     & ResNet-50  & 256  & 200  & 69.7 \\
  BYOL~\cite{albrecht2020}     & ResNet-50  & 4096  & 200  & \underline{70.6} \\
  SimSiam~\cite{chen2021exploring}     & ResNet-50  & 256  & 200  & 70.0 \\
  ProPos (\textbf{ours})           & ResNet-50 & 256  & 200  & \textbf{72.2}    \\
  \midrule
  CPC~\cite{oord2018representation} & ResNet-101& 512  & -    & 48.7 \\
  SeLa~\cite{asano2019self} & ResNet-50 & 1024 & 400  & 61.5 \\
  PIRL~\cite{misra2020self} & ResNet-50 & 1024 & 800  & 63.6 \\
  SimCLR~\cite{chen2020simple}   & ResNet-50 & 4096 & 1000 & 69.3 \\
  BYOL~\cite{albrecht2020} & ResNet-50 & 4096 & 1000 & 74.3 \\
  SwAV~\cite{caron2020unsupervised} & ResNet-50 & 4096 & 800  & 75.3 \\
  \shline
  \end{tabular*}
\end{table}

\begin{table*}[t]
  \centering
  \caption{
    Ablation studies~(NMI/ACC/ARI) for different self-supervised learning frameworks, positive sampling alignment (PSA), and prototype scattering loss (\lossname) for \methodname. The best and second best results are shown in bold and underline, respectively.
  }
  \label{tab:ablation_study}

  \begin{tabular*}{1\linewidth}{@{\extracolsep{\fill}}lcccccc}
  \shline
  \multirow{2}{*}{\textbf{Method}} & \multicolumn{3}{c}{\textbf{CIFAR-10}} & \multicolumn{3}{c}{\textbf{CIFAR-20}} \\
  \cmidrule{2-7}
    & NMI & ACC & ARI & NMI & ACC & ARI \\ \midrule
  CC~\cite{li2021contrastive}  & 66.1\std{0.3} & 74.6\std{0.3} & 58.3\std{0.4} & 46.4\std{0.3} & 45.0\std{0.1} & 29.5\std{0.2} \\
  CC +  \lossname & 74.3\std{0.4} & 83.4\std{0.5} & 69.6\std{1.0} & 48.3\std{0.2} & 49.1\std{0.2} & 32.2\std{0.4} \\
  PCL~\cite{li2020prototypical} & 77.6\std{0.1} & 85.5\std{0.1} &73.4\std{0.0} & 50.0\std{0.3} &48.6\std{0.7} &32.7\std{0.4} \\
  BYOL~\cite{grill2020bootstrap} & 79.4\std{1.7} & 87.8\std{1.7} & 76.6\std{2.8} & 55.5\std{0.6} & 53.9\std{1.6} & 37.6\std{0.9} \\
  BYOL + CC & 76.6\std{3.1} & 86.3\std{2.7} & 73.8\std{4.7} & 51.0\std{2.0} & 48.9\std{3.0} & 33.3\std{2.9} \\
  BYOL + PCL & 74.4\std{2.3} & 85.3\std{0.9} & 71.4\std{1.4} & 49.7\std{0.7} & 46.9\std{0.7} & 27.8\std{1.5} \\
  \midrule
  \methodname (\textbf{ours}) & \underline{85.1}\std{0.5} & \underline{91.6}\std{0.4} & \underline{83.5}\std{0.7} & \textbf{58.2}\std{0.3} & \textbf{57.8}\std{0.2} & \textbf{42.3}\std{0.3} \\
  \methodname w/o \lossname & 79.4\std{0.9} & 87.9\std{0.5} & 76.4\std{1.1} & 57.0\std{0.0} & 55.0\std{0.6} & 39.8\std{1.1} \\
  \methodname w/o PSA & 83.4\std{1.2} & 90.3\std{0.9} & 81.1\std{1.7} & 56.6\std{0.4} & 55.1\std{0.5} & 40.7\std{1.0} \\
  \methodname w/o PSL-uniformity & 79.6\std{0.7} & 87.8\std{1.5} & 76.5\std{2.1} & 56.7\std{0.3} & 56.6\std{1.4} & 39.7\std{1.1} \\
  \methodname w/o PSL-alignment & \textbf{85.3}\std{0.2} & \textbf{92.1}\std{0.1} & \textbf{84.4}\std{0.3} & \underline{57.2}\std{0.3} & \underline{57.3}\std{0.6} & \underline{41.7}\std{0.5} \\
  \shline
  \end{tabular*}
\end{table*}

On the ImageNet-10, our \methodname achieves competitive performance as compared to IDFD~\cite{tao2021clustering} since this dataset is relatively small with only 13K images, which cannot arise discriminative differences for current state-of-the-art methods. On the ImageNet-Dogs, a fine-grained dataset containing different species of dogs from the ImageNet dataset, there are almost 20\% improvements over previous state-of-the-art work. The contrastive-based methods cannot handle this kind of dataset due to severe class collision issue that pushes away the instances from the same class. Meanwhile, IDFD can deal with this problem to some degree thanks to the feature decorrelation along with the instance discrimination.
Furthermore, BYOL and SimSiam can achieve significant improvements, which suggests a great potential for non-contrastive representation learning for deep clustering \emph{without} suffering class collision issue. However, our \methodname has introduced further substantial improvements for deep clustering by addressing existing issues. Although CC~\cite{li2021contrastive} and TCL~\cite{li2022twin} achieve the better NMIs on STL-10, we highlight that they use a large image size of $224\times224$ for all datasets, which is not fair to ours.

\subsubsection{Results on large-scale datasets}

\begin{figure*}[t]
  \centering
  \includegraphics[width=0.96\linewidth]{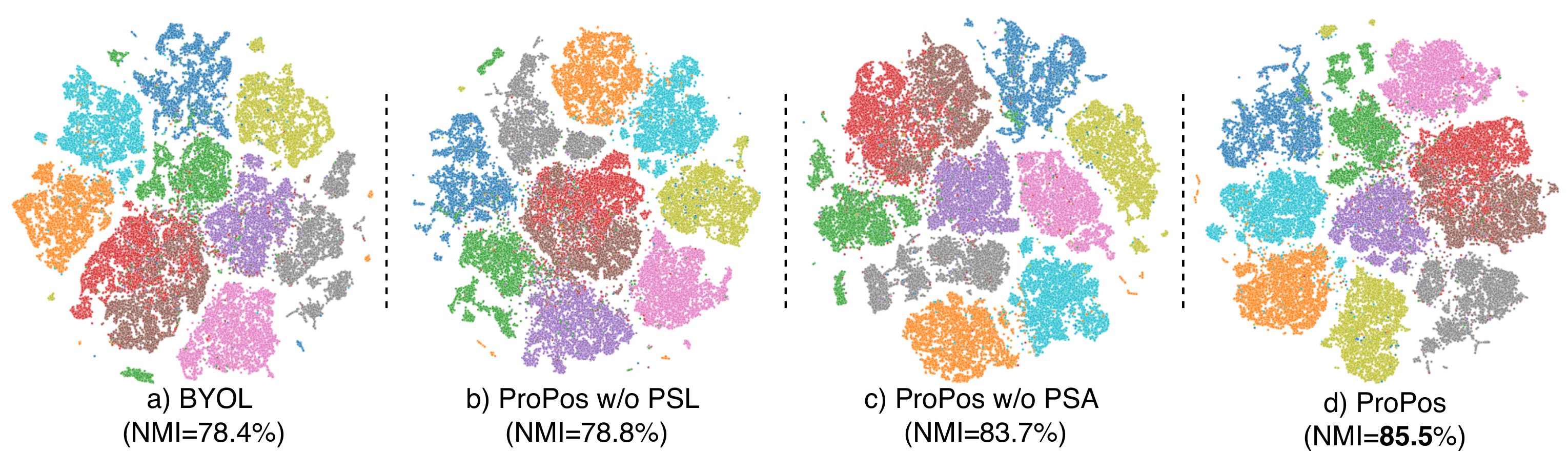}
  \caption{
    Visualization of feature representations learned by different representation learning frameworks and our proposed \methodname on CIFAR-10 with t-SNE. Zoom in for better view.
  }
  \label{fig:clustering_quality_ablation}
\end{figure*}

To validate the effectiveness of our method on large-scale datasets with large number of classes, we evaluate it on Tiny-ImageNet and ImageNet-1k, which contains 200 and 1000 classes, respectively. The results are reported in Tables~\ref{tab:tiny_imagenet} and~\ref{tab:result_imagenet}.
We note that we exclude the methods that their authors did not report results on the corresponding datasets from Tables~\ref{tab:tiny_imagenet} and~\ref{tab:result_imagenet}.
For ImageNet-1k, we strictly follow the settings in~\cite{li2020prototypical} and employed Adjusted Mutual Information~(AMI) to evaluate the performance, the results in Table~\ref{tab:result_imagenet} are referred from~\cite{li2020prototypical}, and we trained a ResNet-50 for 200 epochs same as~\cite{li2020prototypical}. The results show the strong generalization ability of \methodname on complex datasets with a large number of clusters.

To further show the effectiveness of \methodname for downstream classification task, we conducted the linear evaluation on ImageNet-1k dataset, and provide the comparisons with the recent state-of-the-art methods in Table~\ref{tab:imagenet_linear_classifier}. Following the same settings in~\cite{chen2021exploring}, we have trained the linear classifier for 90 epochs with a batch size of 4,096, an initial learning rate of 1.6, cosine learning rate decay, and the SGD optimizer of momentum 0.9 and weight decay 0. Under fair conditions, \methodname outperforms other competitors by a clear margin.

\subsection{Ablation Study}
\label{sec:ablation_study}
Here, we perform detailed ablation studies with both quantitative and qualitative comparisons to provide more insights into why \methodname performs well for deep clustering.

\subsubsection{Quantitative ablation study} 

The quantitative results are shown in Table~\ref{tab:ablation_study}.

\noindent\textbf{Ablation study of PSL and PSA.}\quad
\methodname w/o PSA improves the baseline results by a large margin while \methodname w/o \lossname achieves marginal improvements, which indicates that \lossname is the key to boosting the clustering performance. However, PSA plays an important role in two parts. First, simply using the PSA can stable and further improve the performance than the baseline BYOL~(only instance alignment loss), especially when the number of semantic classes increases for CIFAR-20. Second, PSA can make \methodname better for clustering when \lossname is used in conjunction with the PSA to pull together the neighbor examples. This is because the well-seperated clusters by \lossname can further ensure that PSA samples the positive neighbors that are in the same semantic classes than the one without \lossname.
On the other hand, \lossname only considers inter-cluster distance, and cannot benefit within-cluster compactness. Therefore, the combination of the positive sampling and \lossname achieves the best clustering results, where \lossname aligns the cluster centers between two augmented views and maximizes the inter-cluster distance, and PSA improves the within-cluster compactness.

To further explore the effect of \lossname, following Eq.~(\ref{eq:decoupled_proto_contr}) we split \lossname into alignment and uniformity terms denoted as \lossname-alignment and \lossname-uniformity in Table~\ref{tab:ablation_study}. It is clear that the performance gain from the alignment term is marginal while the gain from the uniformity term is significant. For only alignment term, we compute the loss after predictor network instead of feature extractor, otherwise, representation collapse will turn out. This indicates that uniformity is more important than alignment which can scatter the prototypes to encourage the uniform representations to address the issues in Sec.~\ref{sec:justification}. However, the alignment term is essential to stabilize the training process, as demonstrated in the results for CIFAR-20 with more clusters.

\noindent\textbf{Ablation study of self-supervised learning framework.}\quad
To alleviate the bias of self-supervised learning framework, we conduct experiments in two folds. First, we integrate CC~\cite{li2021contrastive} and PCL~\cite{li2020prototypical} into BYOL; the results in Table~\ref{tab:ablation_study} show that both of them compromise clustering performance and become unstable. This is because CC contrasts the cluster probability not helpful for representation learning, and PCL would collapse without negative examples; see supplemental Fig.~\ref{fig:pcl_moco_and_byol} for detailed analysis.
We emphasize that BYOL has addressed the class collision issue. Instead, this paper proposes the \methodname with \lossname to scatter the prototypes by maximizing inter-cluster distance and positive sampling to improve within-cluster compactness.
Second, we replace the cluster head of CC with our \lossname on the representations while keeping other hyper-parameters unchanged. Although class collision issue remains, the significant improvements over CC on both datasets suggest that 1) \lossname over representation prototypes is better than the one over cluster probabilities, and 2) \lossname can be generalized to other self-supervised learning frameworks.

\subsubsection{Qualitative ablation study} 
Fig.~\ref{fig:clustering_quality_ablation} visualizes the distribution of representations learned from BYOL, \methodname w/o \lossname, \methodname w/o PSA, and our \methodname.
\methodname w/o PSA leverages \lossname to discriminate different clusters by maximizing the inter-cluster distance, and produces more uniform representations.
Although \methodname w/o \lossname with only PSA achieves marginal improvements and does not produce a significant difference than BYOL, the positive sampling can further improve the within-cluster compactness with only \lossname via the sampling-based instance alignment loss to pull together the neighbor samples.

\subsubsection{Effect of predefined number of clusters}
\begin{figure}[t]
  \centering
  \includegraphics[width=0.9\linewidth]{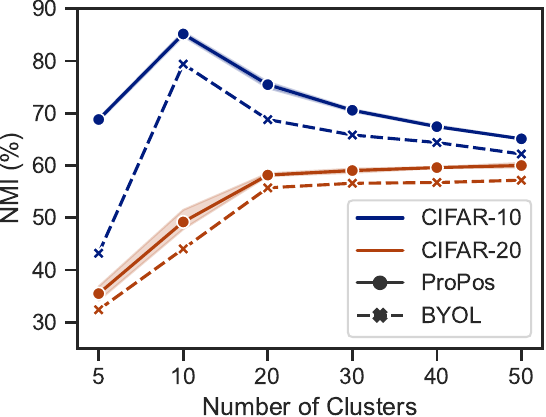}
  \caption{
    The effect of the predefined number of clusters $K$ on CIFAR-10/20 datasets.
  }
  \label{fig:overclustering}
\end{figure}

In the above experiments, the number of clusters is predefined as the number of ground-truth classes, which cannot be identified in the practical scenarios.
To this end, we conduct experiments on CIFAR-10 and CIFAR-20 with different number of clusters, \ie, $K\in \{5, 10, 20, 30, 40, 50\}$. We reported NMIs following~\cite{caron2018deep} in Fig.~\ref{fig:overclustering}. We note that the predefined $K$ of \methodname during the training of \methodname is the same as the $K$ in $k$-means clustering process for evaluation. To further investigate the influences of $K$, we also reported the results of vanilla BYOL during $k$-means clustering.

The results in Fig.~\ref{fig:overclustering} demonstrate that BYOL and \methodname have the same behavior on these two datasets.
For over-clustering cases~($K$ is larger than the true number of classes), the trends on these two datasets are opposite.
Specifically, over-clustering leads to a performance drop for CIFAR-10, but it leads to an increase for CIFAR-20. However, our \methodname can still produce large improvements over BYOL with the same predefined number of clusters.
The opposite results are due to the significant difference between these two datasets. Although having the same number of samples, CIFAR-10 has 10 distinct classes while CIFAR-20 has, in fact, 100 classes but uses 20 super-classes instead. Same trends are also consistently reported in~\cite{caron2018deep}.
In other cases, if the representations are well aligned within the same semantic clusters, the over-clustering would try to destroy the structures of the clusters and push the semantically similar examples away, which certainly compromises the clustering performance.
For under-clustering cases, the clustering performance has been significantly harmed for both two datasets and two methods. In supplemental Fig.~\ref{fig:underclustering_vis}, we have visualized the learned representations for under-clustering, which shows that the examples from the same semantic classes can be still clustered together.

\subsection{Hyperparameter Analysis}
\label{sec:hyperparameter_analysis}
\begin{figure*}[h]
  \centering
  \includegraphics[width=0.9\linewidth]{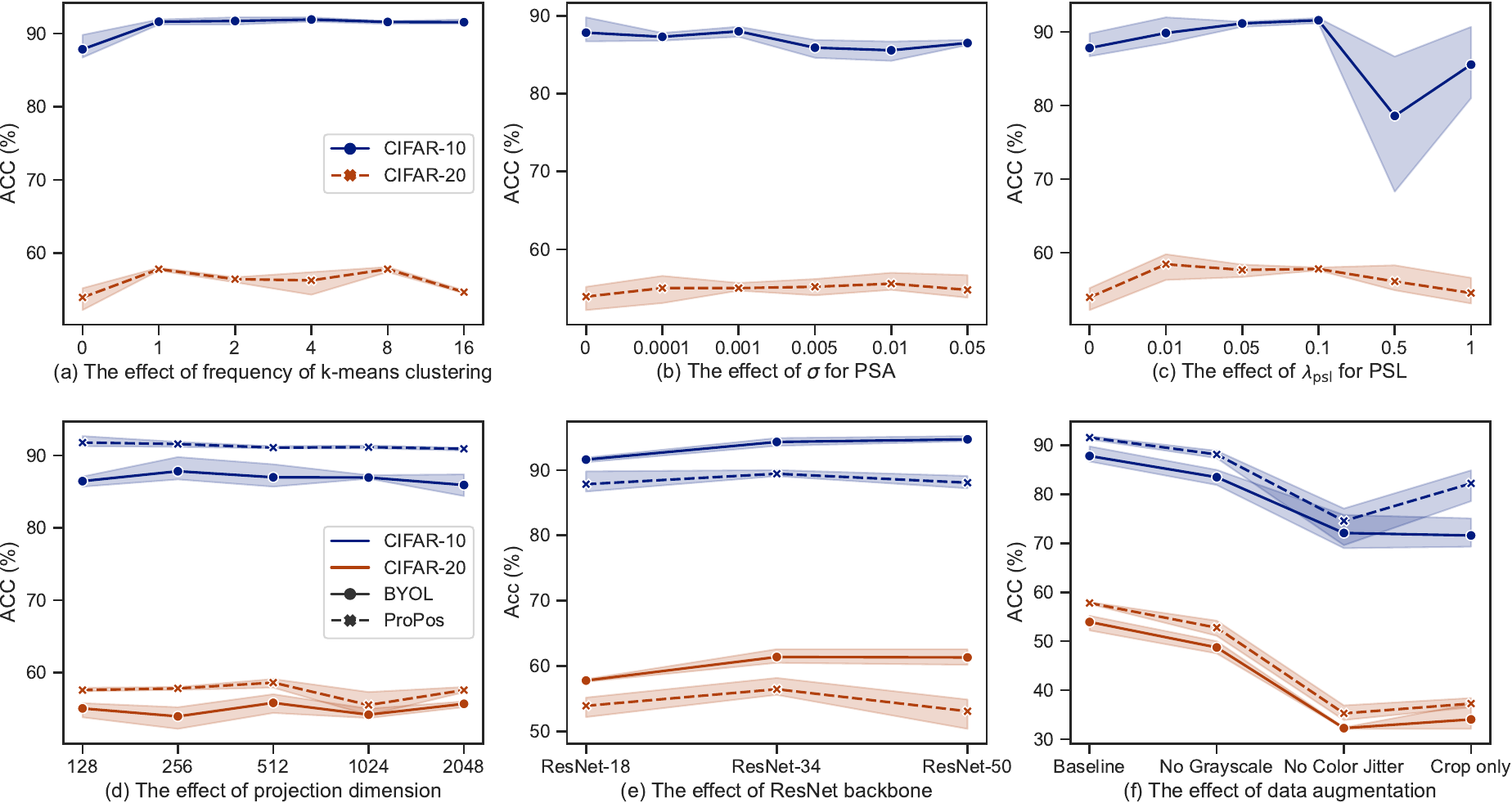}
  \caption{
    Effects of different hyperparameters in \methodname.
  }
  \label{fig:hyperparameter_analysis}
\end{figure*}

To investigate the effect of different hyperparameters, we conduct extensive experiments on the CIFAR-10/20 datasets. For the projection dimension, backbone, and data augmentation of SSL, we adopted the BYOL as the baseline method. The results are reported in Fig.~\ref{fig:hyperparameter_analysis}.
\begin{table*}[t]
    \centering
    \caption{The training time~(hours) in the settings of 1,000 epochs, 4 V100 GPUs, and ResNet-50.}
    \label{tab:training_time}
    {
    \begin{tabular*}{1\linewidth}{@{\extracolsep{\fill}}lcccccc}
    \shline
    \textbf{Method} & \multicolumn{1}{c}{\textbf{CIFAR-10}} & \multicolumn{1}{c}{\textbf{CIFAR-20}} & \multicolumn{1}{c}{\textbf{STL-10}} & \multicolumn{1}{c}{\textbf{ImageNet-10}} & \multicolumn{1}{c}{\textbf{ImageNet-dogs}} & \multicolumn{1}{c}{\textbf{Tiny-ImageNet}} \\
    \midrule
    BYOL~\cite{grill2020bootstrap} & 9.0 & 9.0 & 14.7 & 1.7 & 2.6 & 13.0 \\
    \methodname~($r=1$)   & 10.9(+1.9) & 11.0(+2.0) & 15.8(+1.1) & 2.7(+1.0)  & 3.7(+1.1) & 15.7(+2.7)  \\
    \shline
    \end{tabular*}
    }
\end{table*}

\subsubsection{Frequency of performing $k$-means clustering} 
\methodname performs $k$-means clustering for every $r$ epoch. Here, we study how different $r$ influences the clustering performance. The results in Fig.~\ref{fig:hyperparameter_analysis}(a) demonstrate \methodname is robust to large $r$ and the cluster pseudo-labels, which means it is not necessary to perform clustering for every epoch so that the computation cost can be significantly reduced. In summary, we suggest that $r$ can be set to $[1, 8]$ by considering the datasets and computation resources.

\subsubsection{The hyperparameter $\sigma$ in PSA} 
The hyperparameter $\sigma$ in PSA controls the degree of positive sampling. Taking a look at $\sigma\sim [0, 10^{-3}]$ in Fig.~\ref{fig:hyperparameter_analysis}(b), although introducing the positive sampling into BYOL causes a slight drop on CIFAR-10, the clustering performance becomes more stable as evidenced by the standard deviation. This is because the neighbors of one sample are regarded as positive examples. Besides, the performance for CIFAR-20 has increased over baseline with the standard deviation reduced. These results indicate that positive sampling can improve the stability of performance. However, when $\sigma$ is too large, the performance becomes unstable and drops a lot. It is not surprising since during positive sampling with large $\sigma$, the instances from other clusters could be sampled and regarded as positive examples. Therefore, we suggest setting $\sigma$  to a small value, saying $(0, 10^{-3}]$.

\subsubsection{The hyperparameter $\lambda_{\mathrm{psl}}$ for \lossname} 
The hyperparameter $\lambda_{\mathrm{psl}}$ controls the importantance of \lossname. The results in Fig.~\ref{fig:hyperparameter_analysis}(c) suggest that \methodname is robust to different choices on CIFAR-20. However, the higher $\lambda_{\mathrm{psl}}$ leads to instability on CIFAR-10. The possible reason is that CIFAR-20 is more diverse and has more semantic classes than CIFAR-10~(100 versus 10). Therefore, we suggest that $\lambda_{\mathrm{psl}}$ can be set to $[0.01, 0.1]$, which has demonstrated superior performance on both two datasets.

\subsubsection{Projection dimension} 
The projection dimension describes the embedding space of SSL. The results in Fig.~\ref{fig:hyperparameter_analysis}(d) shows that \methodname achieves consistent and significant performance improvement over baseline regardless of different projection dimension.

\begin{table}[t]
    \centering
    \caption{
        Clustering results (\%) on the subsets of ImageNet. 
    }
    \label{tab:scan_imagenet_subset}
    \begin{tabular*}{1\linewidth}{@{\extracolsep{\fill}}lcccccc}
        \shline
        \textbf{ImageNet} & \multicolumn{2}{c}{\textbf{50 Classes}} & \multicolumn{2}{c}{\textbf{100 Classes}} & \multicolumn{2}{c}{\textbf{200 Classes}} \\
        \cmidrule{2-7}
        Method   & NMI            & ARI           & NMI            & ARI            & NMI            & ARI            \\
        \midrule
        \tabincell{l}{$k$-means w/ \\ pre-trained MoCo}  & 77.5           & 57.9          & 76.1           & 50.8           & 75.5           & 43.2           \\
        \tabincell{l}{SCAN~\cite{van2020scan} after\\ clustering
step}     & 80.5           & 63.5          & 78.7           & 54.4           & 75.7           & 44.1           \\
        \tabincell{l}{SCAN~\cite{van2020scan} after\\ self-labeling
step}     & \underline{82.2} & \underline{66.1} & \underline{80.8} & \underline{57.6} & \underline{77.2} & \underline{47.0} \\
        \methodname~(\textbf{ours})   & \textbf{82.8}  & \textbf{69.1} & \textbf{83.5}  & \textbf{63.5}  & \textbf{80.6}  & \textbf{53.8}  \\
        \shline
    \end{tabular*}
\end{table}
\subsubsection{ResNet backbone}\quad 
Fig.~\ref{fig:hyperparameter_analysis}(e) shows that with the deeper ResNet networks, \methodname achieves significant improvements with small standard deviations for clustering, demonstrating its superior stability and performance against the baseline method.

\begin{table*}[t]
    \centering
    \caption{
        Clustering results (\%) on long-tailed datasets of different self-supervised learning frameworks and our proposed \methodname. 
    }
    \label{tab:long_tailed_datasets}
    \begin{tabular*}{1\linewidth}{@{\extracolsep{\fill}}l*{6}{c}}
    \shline
    \multirow{3}{*}{\textbf{Method}} & \multicolumn{3}{c}{\textbf{CIFAR-10-LT}}  & \multicolumn{3}{c}{\textbf{CIFAR-20-LT}} \\
    \cmidrule{2-7}
     & \multicolumn{1}{c}{NMI} & \multicolumn{1}{c}{ACC} & \multicolumn{1}{c}{ARI} & \multicolumn{1}{c}{NMI} & \multicolumn{1}{c}{ACC}      & \multicolumn{1}{c}{ARI} \\
    \midrule
    MoCo v2~\cite{he2020momentum} & 46.7\std{0.1} & 33.4\std{0.3} & 27.7\std{0.0} & 31.2\std{0.3} & 28.2\std{0.2} & 16.1\std{0.3} \\
    BYOL~\cite{grill2020bootstrap} & 51.6\std{1.0} & 41.3\std{0.4} & 30.8\std{0.4} & 41.9\std{0.4} & 34.6\std{0.5} & 22.3\std{1.0} \\ \hline
    \methodname w/o \lossname & \underline{53.1}\std{0.7} & \underline{42.7}\std{0.4} & \underline{31.6}\std{0.8} & \underline{43.4}\std{0.8} & \underline{35.1}\std{0.6} & \underline{24.0}\std{0.1} \\
    \methodname (\textbf{ours}) & \textbf{55.3}\std{0.4} & \textbf{43.9}\std{0.1} & \textbf{36.3}\std{0.3} & \textbf{44.6}\std{0.2} & \textbf{39.0}\std{0.7} & \textbf{27.3}\std{0.2} \\
    \shline
    \end{tabular*}
\end{table*}

\subsubsection{Data augmentation} 
Data augmentation is important for self-supervised learning. Fig.~\ref{fig:hyperparameter_analysis}(f) shows that the performance drops for both BYOL and \methodname when removing some data augmentations. However, the results suggest that \methodname still performs more stable and is robust to data augmentations.

\subsection{Computational Cost}\label{sec:computational_cost}

The main additional computational cost of \methodname is the $k$-means clustering procedure. We have implemented the $k$-means algorithm with $k$-means++~\cite{arthur2006k} initialization using PyTorch to utilize the GPU and accelerate the clustering process. We performed $k$-means clustering with 10 different initialization in cosine distance. Table~\ref{tab:training_time} summarizes the training time of \methodname and BYOL on different datasets. \methodname does not introduce much additional computational cost.
Besides, as suggested in the results of Fig.~\ref{fig:hyperparameter_analysis}(a), \methodname is robust to the different $r$, so there is no need to perform $k$-means for every epoch. Therefore, the training time can be further reduced for \methodname. The computational cost of the \lossname is also small since we build the cluster centers within mini-batch, saying that \methodname does not need additional memory to store the cluster centers. Consequently, considering the promising performance improvements, the additional computational cost is relatively affordable.

\subsection{Subsets of ImageNet}

In addition, we also reported the clustering results on ImageNet subsets like SCAN~\cite{van2020scan} in Table~\ref{tab:scan_imagenet_subset}.
We strictly follow the settings in~\cite{van2020scan}: we have adopted the same 50, 100, and 200 classes from ImageNet, clustered on the training set, and tested on the validation set. We have used the same experimental settings as the other benchmarked datasets and trained \methodname with ResNet-50 for 300 epochs. We note that SCAN has used the pre-trained model of MoCo trained on the full ImageNet for 800 epochs. The results are directly referred from their published paper including $k$-means with pre-trained MoCo, SCAN after the clustering step, and SCAN after the self-labeling step. With much fewer training epochs and training data, \methodname still produces better performance by a clear margin, demonstrating the superiority of \methodname.

\subsection{Long-tail Dataset}
We also conducted additional experiments in Table~\ref{tab:long_tailed_datasets} to demonstrate the ability of \methodname handling the long-tailed datasets.
We built the long-tailed version of CIFAR-10 and CIFAR-20, termed CIFAR-10-LT and CIFAR-20-LT using the codes of~\cite{tang2020long}, which follows~\cite{zhou2020bbn,cao2019learning}. Specifically, they were built upon the training datasets under the control of data imbalance ratio $\min(\{N_k\}_{k=1}^K)/\max(\{N_k\}_{k=1}^K)=0.1$. Consequently, the samples in the long-tailed datasets are almost all in the minority classes~(head), versus few samples in other classes~(tail). MoCo v2 cannot handle this problem well due to the class collision issue, as a result, the samples in the head will be pushed away and the ones in the tail will be mixed together. BYOL and \methodname do not need negative examples so that they outperform MoCo v2 by a large margin. By introducing 
PSA and \lossname, we can further boost the clustering performance of BYOL.

\subsection{Boosting Performance with Memory Queue}
To better represent the prototype of one class, the mini-batch should contain a sufficient number of samples. However, this would significantly increase the requirement of GPU memory when one dataset has a large number of clusters. Here, we highlight that although we use a mini-batch size of $256$, we find that \methodname generalizes well on the datasets such as Tiny-ImageNet with 200 classes (about 1 sample per class in a mini-batch) and ImageNet-1k with 1,000 classes (about 0.25 sample per class in a mini-batch).

\begin{figure}[t]
  \centering
  \includegraphics[width=1\linewidth]{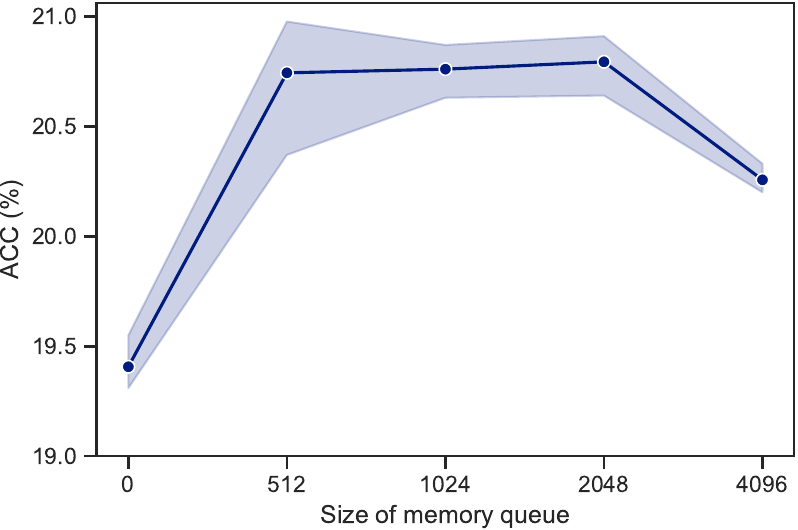}
  \caption{
    The performance of \methodname using a memory queue with different sizes on Tiny-ImageNet.
    We repeated each run for 3 times and reported the mean and std values.
    }
  \label{fig:memory_queue}
\end{figure}
To compute class prototypes accurately with a small mini-batch size, we propose to employ a memory queue for updating prototypes. In the vanilla \lossname, we use the representations within mini-batch $\mathcal{B}$ to estimate the prototypes in Eqs.~\eqref{eq:compute_centers_1} and~\eqref{eq:compute_centers_2}.
With a memory queue $\mathcal{Q}$ used in~\cite{he2020momentum} to store the representations from the momentum-updated encoder, we can update the prototypes with samples in both mini-batch and memory queue, $\mathcal{B} \cup \mathcal{Q}$.

To show the effectiveness of the memory queue, we conduct the experiments on Tiny-ImageNet as it is challenging enough with 200 classes, and the corresponding results are shown in Fig.~\ref{fig:memory_queue}; we trained the model with the same settings as detailed in Section~\ref{sec:implementation_details} except for 200 epochs.
As shown in Fig.~\ref{fig:memory_queue}, the performance can be further improved with more samples used to compute the prototypes.
However, the performance could drop when the size of memory queue becomes too large, saying 4,096 in Fig.~\ref{fig:memory_queue}. The reason is that when the memory queue is too large, the representations enqueued at early iterations may be far away from the true ones due to the longer update of the encoder~\cite{he2020momentum}.
Moreover, using the memory queue brings more computational costs and introduces an additional hyperparameter---the size of memory queue.
Therefore, \methodname works well when the mini-batch size is small compared to the number of classes, and its performance can be further improved with a memory queue.
\section{Discussion}\label{sec:discuss}

In this section, we discuss the differences between our method and previous works.
\subsection{Relation to CC} 
Although both \lossname and CC~\cite{li2021contrastive} are class-level contrastive loss, which perform contrastive learning at the cluster level, they have the following difference.
\begin{itemize}
\item The class-level contrastive loss in CC implements the contrastive loss on \emph{the cluster probabilities} while ours on \emph{the representation of cluster centers}. Implementing contrastive loss on the cluster probability in~\cite{li2021contrastive} would lose the semantic information of the learned representations, which is not helpful for representation learning. Specifically, given the $\vct{x} \in \mathcal{B}$, CC obtains the cluster assignments $\mat{P}_k=[p(k \vert \boldsymbol{x}^{(1)}), \ldots, p(k \vert \boldsymbol{x}^{(N)})]$ from one view and $\boldsymbol{P}_{k}{'}$ from another view, and then contrasts $\boldsymbol{P}_k$ and $\boldsymbol{P}_{k}{'}$ at the cluster level using the InfoNCE loss. In contrast, \lossname implements the contrastive loss on the representation of the cluster centers within a mini-batch using the pseudo-labels from $k$-means clustering. As a result, \lossname is able to sense the semantic information of the latent space and make the representations of clusters more discriminative and suitable for the clustering task.
\item The class-level contrastive loss in CC does not encourage cluster uniformity while our \lossname does. CC still needs the instance-wise contrastive loss to encourage instance uniformity, which inevitably introduces the class collision issue. 
\end{itemize}

In addition to these differences, experimentally, we also compare the PSL and CC~\cite{li2021contrastive} in the same BYOL framework and report the results in Table~\ref{tab:ablation_study}. The experimental results show that compared to BYOL, BYOL+CC drops the performance and makes the training unstable as CC fails to encourage uniform representations. Under the same conditions, \methodname achieves significant improvements over BYOL+CC.

\subsection{Relation to GCC and WCL} To alliterate the class collision issue, 
GCC~\cite{zhong2021graph} and WCL~\cite{zheng2021weakly} built a graph to label the neighbor samples as pseudo-positive examples. Then, they enforce the two data augmentations of one example to be close to its multiple pseudo-positive examples using a supervised contrastive loss. GCC adopted a moving-averaged memory bank for the graph-based pseudo-labeling while WCL built the graph within a mini-batch. GCC and WCL mainly focus on how to effectively select positive examples from mini-batch/memory bank to alleviate the class collision issue. Here, we divide the class collision issue into the following two cases:
\begin{itemize}
\item Negative class collision issue: negative examples may not be \emph{truly negative}, which is the case contrastive learning faces.
\item Positive class collision issue: positive examples may not be \emph{truly positive}, which is a new case raised in GCC and WCL.
\end{itemize}

Consequently, they still suffer from the positive class collision issue as the selected pseudo-positive examples may not be truly positive. In addition to this, they also suffer from the negative class collision issue since they still need negative examples for instance-wise contrastive learning.

We summarize the difference between our PSA and theirs in the following four aspects.
\begin{itemize}
\item GCC and WCL select the examples that exist in the dataset (mini-batch/memory bank) while ours samples examples from the latent space that may not exist in the dataset.

\item GCC and WCL select neighbor examples in a graph as pseudo-positive examples that may not be \emph{truly positive} while ours samples examples around the instance in the embedding space that can be assumed to be \emph{truly positive}.

\item GCC and WCL still rely on instance-wise contrastive loss that could lead to class collision issue while ours can avoid class collision issue by using non-contrastive BYOL.

\item GCC and WCL require additional computational cost for graph construction while ours is rather cheap in sampling one example in the embedding space.

\end{itemize}

\subsection{Relation to PCL}
Here, we summarize the difference between our \methodname and PCL~\cite{li2020prototypical} in terms of the losses and EM frameworks.
First, we summarize the difference between our \lossname and ProtoNCE loss used in PCL as follows.
\begin{itemize}
\item Our \methodname can avoid class collision issue while PCL cannot. \methodname is based on BYOL that does not require negative examples for representation learning while PCL is based on instance-wise contrastive loss that requires a number of negative examples for representation learning, inevitably leading to class collision issue.

\item The proposed \lossname in \methodname is conceptually different from the ProtoNCE in PCL. \lossname is to maximize the inter-cluster distance to form a uniformly distributed space while ProtoNCE is to minimize the instance-to-cluster distance to improve the within-cluster compactness. The within-cluster compactness of \methodname is improved by the proposed positive sampling alignment.

\item Pure \lossname can work well for deep clustering while ProtoNCE requires another InfoNCE to form uniformly distributed space. This is a direct result of the different designs of the losses. \lossname can maximize the inter-cluster distance to form a uniformly distributed space while ProtoNCE suffers from collapse without the help of another InfoNCE to form such a space.
\end{itemize}

Second, we summarize the difference between our \methodname and PCL in the EM framework. Formulating \methodname into an EM framework can offer more insights about \methodname and make it easy to understand.
Although both in an EM framework, the M-step in PCL is significantly different from the one in our \methodname. More specifically, the M-step in PCL is to optimize the ProtoNCE, which is an \emph{instance-to-prototypes contrastive loss} to improve the within-cluster compactness while the M-step in our \methodname is to optimize the proposed \lossname, which is a \emph{prototypes-to-prototypes contrastive loss} to maximize the inter-cluster distance for better clustering performance. In addition, \methodname also proposes a positive sampling alignment by sampling positive examples around each sample to improve within-cluster compactness.

\subsection{Relation to Instance-reweighted Contrastive Loss}
We provide a new perspective to understand the proposed cluster-wise \lossname from instance-reweighted contrastive loss~\cite{mitrovic2020representation}.
Here, we first focus on analyzing the alignment term of \lossname. By substituting  $\vct{\mu}_{k}$ and $\vct{\mu}_{k}^\prime$ into the prototypical alignment term of \lossname, we can rewrite the alignment term as:
\begin{align}
  &\frac{1}{K} \sum_{k=1}^K - \frac{\vct{\mu}_k\T\vct{\mu}_k^\prime}{\tau}\\
  =& \frac{1}{K} \sum_{k=1}^K - \frac{1}{\tau} \frac{\sum_{\mat{x} \in \mathcal{B} }  p(k\vert \mat{x}) f(\mat{x})\T}{c_k} \frac{\sum_{\mat{x} \in \mathcal{B}}  p(k\vert\mat{x}) f^\prime(\mat{x})}{c_k^\prime}\\
  =& \frac{1}{K} \sum_{k=1}^K - \frac{1}{\tau} \frac{\sum_{i=1}^N\sum_{j=1}^N p(k\vert \mat{x}_i) p(k\vert \mat{x}_j) f(\mat{x}_i)\T f^\prime(\mat{x}_j)}{c_k c_k\prime} \\
  =& \sum_{i=1}^N \sum_{j=1}^N - \underbrace{\frac{1}{K} \sum_{k=1}^K \frac{p(k\vert \mat{x}_i) p(k\vert \mat{x}_j)}{c_k c_k^\prime}}_{\vct{w}_{ij}}\frac{f(\mat{x}_i)\T f^\prime(\mat{x}_j)}{\tau},
  \label{eq:reweigted_alignment_psl}
\end{align}
where $p(k\vert \mat{x})\in \{0, 1\}$, $c_k=\|\sum_{\mat{x} \in \mathcal{B} }  p(k\vert \mat{x}) f(\mat{x})\T\|_2$, $c_k^\prime =\|\sum_{\mat{x} \in \mathcal{B}}  p(k\vert\mat{x}) f^\prime(\mat{x})\|_2$, and $\vct{w}=\{\vct{w}_{ij}\}_{i,j=1}^N\in \mathbb{R}^{N\times N}$ denotes the weights of each instance pair. Eq.~\eqref{eq:reweigted_alignment_psl} shows that the alignment term in \lossname can be formulated as an instance-reweighted contrastive loss.

From Eq.~\eqref{eq:reweigted_alignment_psl}, we have the following observation:
\begin{itemize}
    \item When  $\mat{x}_i$ and $\mat{x}_j$ belong to the same cluster, \ie $p(k\vert \mat{x}_i)=p(k\vert \mat{x}_j)=1$, we have $\mat{w}_{ij}>0$. 
    \item When $\mat{x}_i$ and $\mat{x}_j$ belong to the different clusters, we have $\mat{w}_{ij}=0$.  
\end{itemize}
As a result, the alignment term in \lossname only contains the sample pairs belonging to the same clusters, which is similar to supervised contrastive loss~\cite{khosla2020supervised}. Therefore, the alignment term of \lossname is a generalized case of instance-reweighted contrastive loss that takes into account the pseudo-labels.
Similarly, one can observe that the uniformity term in \lossname is to maximize the distance between instances in different clusters ($j\neq k$). 

Therefore, we can understand the proposed \lossname from a perspective of instance-reweighted contrastive loss with cluster labels taken into account.
\section{Conclusion}
\label{sec:conc}

We introduced a novel deep clustering method \methodname, which enjoys the strengths of both contrastive- and non-contrastive-based methods. The proposed positive sampling alignment and prototype scattering loss can lead to within-cluster compactness and well-separated clusters. The results empirically showed that the proposed \methodname outperforms the state-of-the-art methods by a significant margin. Current state-of-the-art methods are mostly beneficial from the progress of self-supervised representation learning while the new trends such as MAE~\cite{he2021masked} have presented more superior performance on downstream tasks, which deserves studying as future work.




\section*{Acknowledgments}
The authors would like to thank the associate editor and two anonymous reviewers for their valuable comments, which greatly improved the quality of this article.

\ifCLASSOPTIONcaptionsoff
  \newpage
\fi



\clearpage
\onecolumn

\begin{center}
\huge
\textbf{
---Supplementary Material---
}
\end{center}

\begin{figure}[h]
    \centering
    \includegraphics[width=1\linewidth]{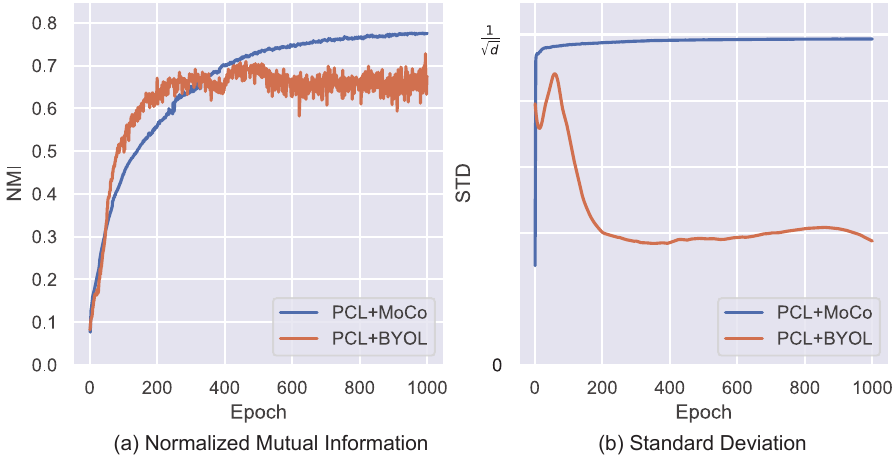}
    \caption{
        Visualizations of NMIs and STDs by applying PCL~\cite{li2020prototypical} to MoCo and BYOL on CIFAR-10.
        Compared to PCL+BYOL, PCL+MoCo performs more stable during clustering with a more uniform distribution of representations.
        The decreasing STDs~(standard deviation of $\ell_2$-normalized features) also indicate that PCL+BYOL suffers from the representation collapse. This is because PCL can only improve the within-cluster compactness. During training, the fixed prototypes of PCL will also be gradually collapsed without negative examples, making the representations for BYOL collapse at the same time.
        These results validate our assumptions that BYOL is not robust to additional clustering losses for clustering tasks, since there is no negative example for BYOL to maintain uniform representations to avoid collapse.
        Our \methodname can avoid the class collision issue and representation collapse by the proposed positive sampling alignment to improve the within-cluster compactness and prototypical scattering loss to maximize the inter-class distance.
    }
    \label{fig:pcl_moco_and_byol}
    
\end{figure}

\begin{figure}[h]
  \centering
  \includegraphics[width=1\linewidth]{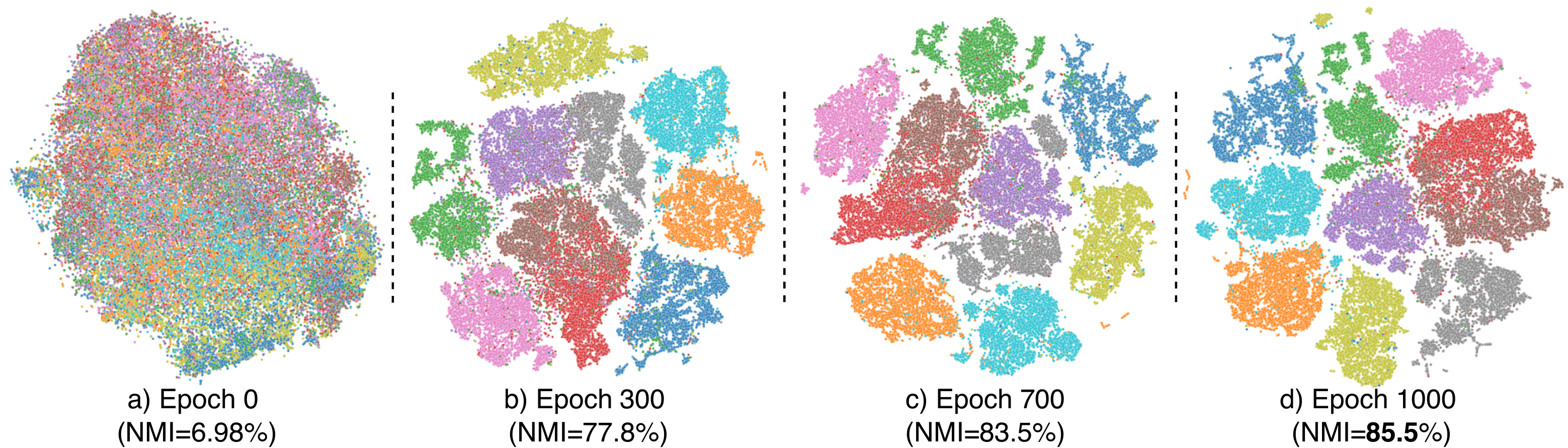}
  \caption{Visualization of feature representations learned by \methodname on CIFAR-10 with t-SNE, for four different training epochs throughout the training process. Different colors denote the different semantic classes. Zoom in for better view. At the beginning, the random-initialized model cannot distinguish the instances from different semantic classes, where all instances are mixed together. As the training process goes, \methodname gradually attracts the instances from the same cluster while pushing the clusters away from each other. Obviously, at the end of the training, \methodname produces clear boundary between clusters and within-cluster compactness.
  }
  \label{fig:clustering_quality}
\end{figure}

\begin{figure}[h]
    \centering
    \includegraphics[width=1\linewidth]{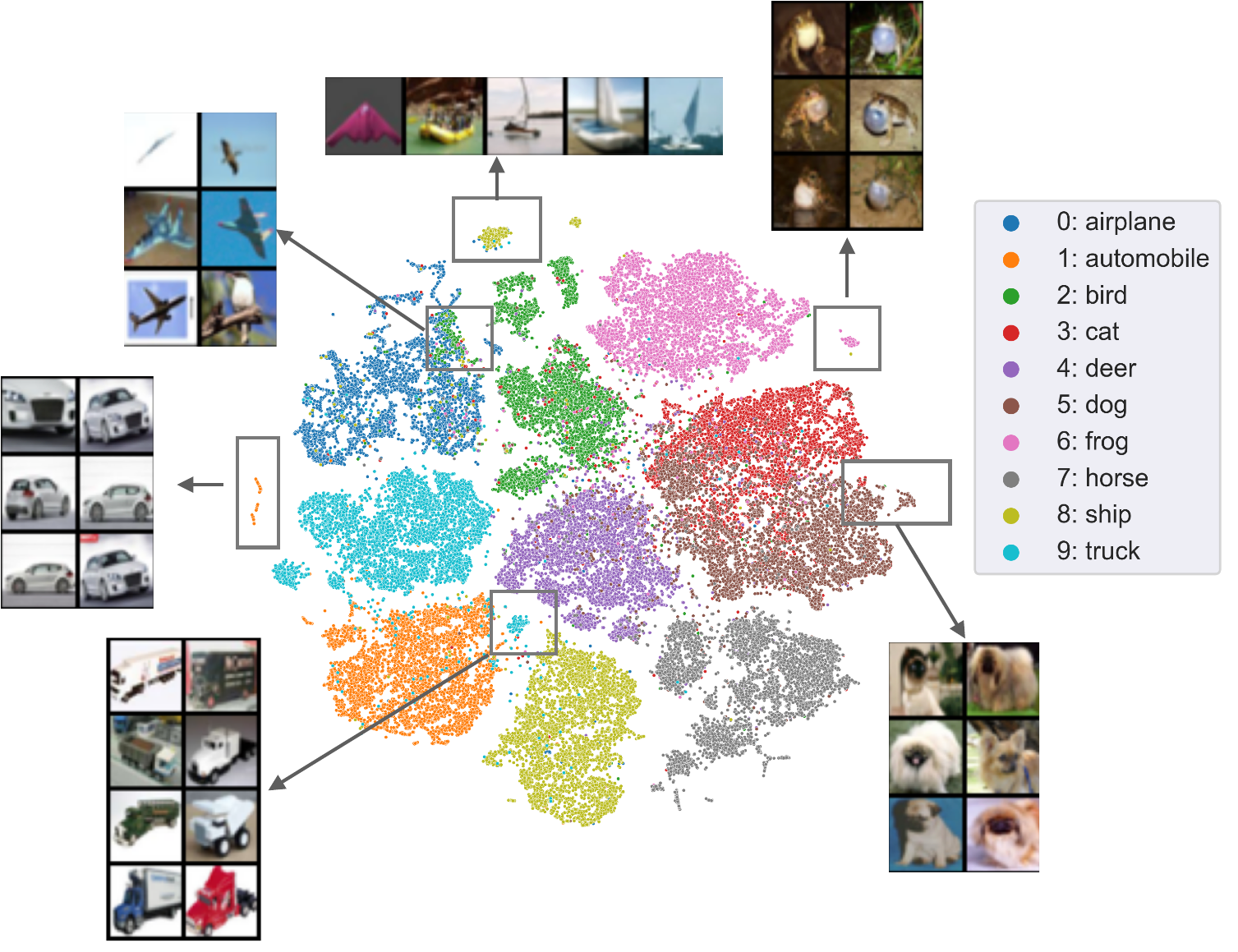}
    \caption{Visualizations for the outlier points produced by the model at 1000-th epoch on CIFAR-10 with t-SNE.}
    \label{fig:outlier_points_vis}
\end{figure}

\begin{figure}[h]
  \centering
  \includegraphics[width=1.0\linewidth]{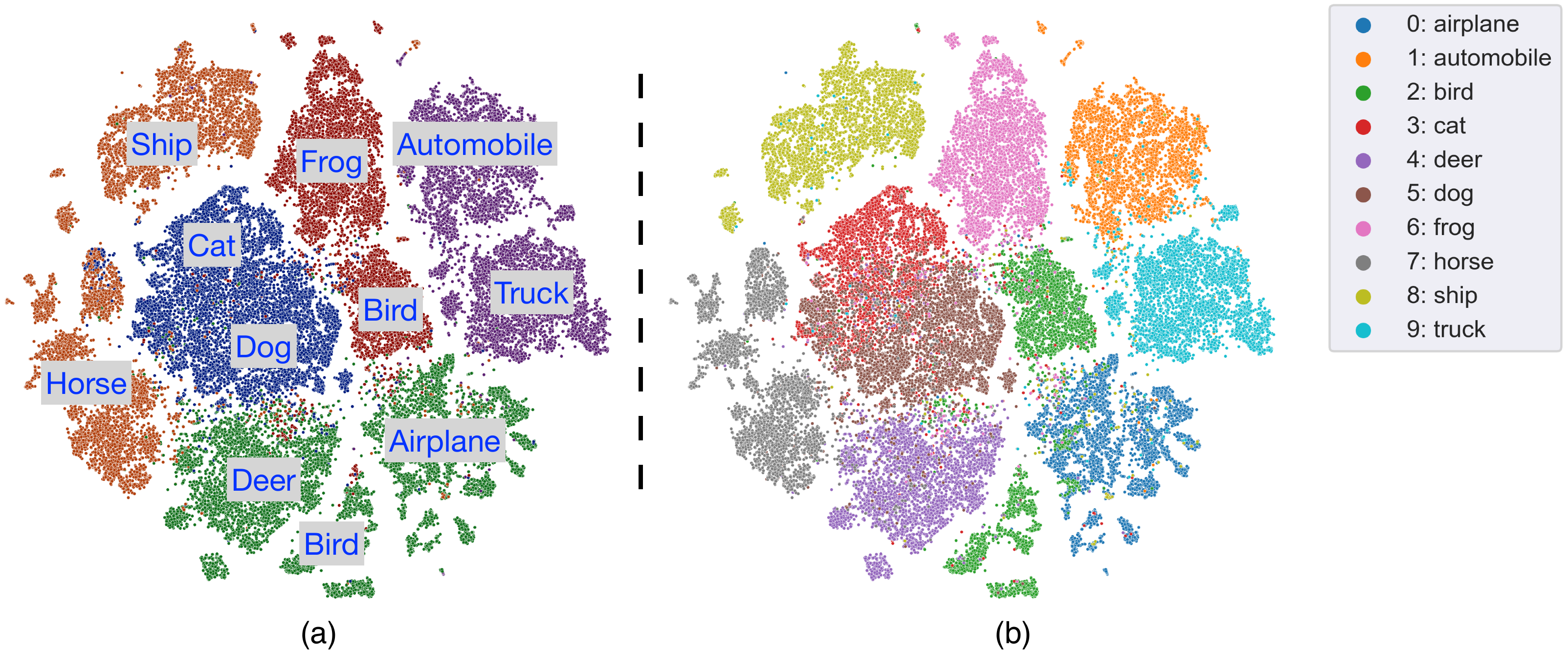}
  \caption{
    Visualization of feature representations learned by \methodname with underestimated $K=5$ on CIFAR-10 with t-SNE: (a) the points are colored according to the pseudo-labels of $k$-means clustering, where the text box denotes the true sematic classes; (b) The points are colored according to the true labels. Zoom in for better view.
  }
  \label{fig:underclustering_vis}
\end{figure}

This supplementary material provides the following extra contents: (1) Sec.~\ref{sec:em_framework} presents detailed derivations of EM framework; (2) Sec.~\ref{sec:addition_results} contains additional experimental results, including Fig.~\ref{fig:pcl_moco_and_byol} for the analysis of applying PCL~\cite{li2020prototypical} to MoCo~\cite{he2020momentum} and BYOL~\cite{grill2020bootstrap}, Fig.~\ref{fig:clustering_quality} for visualization of feature representations throughout the training process, Fig.~\ref{fig:outlier_points_vis} for visualizations for the outlier points, Table~\ref{tab:fair_results_with_imagesize_and_split} for the results under different data split and image size, and Fig.~\ref{fig:underclustering_vis} for visualizations of representations for underclustering.

\section{EM Framework}
\label{sec:em_framework}

In this section, we first describe the \textit{von Mises-Fisher}~(vMF) distribution on the hypersphere, and then derive the evidence lower bound (ELBO) for our expectation–maximization (EM) framework, followed by detailed E-step and M-step. Finally, we describe our proposed prototypical contrastive loss and provide proof for convergence analysis.

\subsection{von Mises-Fisher Distribution} 
Since the features in current SSL methods are usually $\ell_2$-normalized, it is more proper to employ the spherical distribution to describe the features. The \textit{von Mises-Fisher}~(vMF) distribution,  often seen as the Gaussian distribution on a hypersphere, is parameterized by $\vct{\mu}\in \mathbb{R}^{d}$ the mean direction and $\kappa\in \mathbb{R}_{+}$ the concentration around $\vct{\mu}$. For the special case of $\kappa=0$, the vMF distribution represents a uniform distribution on the hypersphere. The probability density function of vMF distribution for the random unit vector $\vct{v}\in \mathbb{R}^{d}$ is defined as:
\begin{align}
    p(\vct{v} \mid \vct{\mu}, \kappa) &= \mathcal{C}_{d}(\kappa) \exp(\kappa \vct{\mu}\T \vct{v}),\quad\text{where}\quad \mathcal{C}_{d}(\kappa) = \frac{\kappa^{d / 2-1}}{(2 \pi)^{d / 2} \mathcal{I}_{d / 2-1}(\kappa)},
\end{align}
where $d$ is the feature dimension, $\|\vct{\mu}\|_{2}=1$, $\mathcal{C}_{d}(\kappa)$ is the normalizing constant, and $\mathcal{I}_v$ denotes the modified Bessel function of the first kind at order $v$. The standard Gaussian distribution $\vct{z}\sim \mathcal{N}(0, \mat{I})$ can be approximately seen as the uniform vMF distribution if the $\vct{z}$ is $\ell_2$-normalized and $d$ is large for the high dimension data.

\subsection{Derivation of Evidence Lower Bound (ELBO)} 
Given the dataset $\mathcal{D}=\{\vct{x}^{(n)}\}_{n=1}^{N}$ with $N$ observed data points that are related to a set of $K$ cluster latent variables, $k\in \mathcal{K}=\{1,2,\ldots, K\}$, the marginal likelihood can be written as:
\begin{align}
    \mathcal{L}(\mathcal{D} ; \mat{\theta}) =\frac{1}{N} \sum_{n=1}^{N} \log p(\mat{x}^{(n)} ; \mat{\theta})=\frac{1}{N} \sum_{n=1}^{N} \log \sum_{k \in \mathcal{K}} p(\mat{x}^{(n)}, k; \vct{\theta}),
    \label{eq:marginal_likelihood}
\end{align}
where $\vct{\theta}$ denotes the model parameters. Eq.~(\ref{eq:marginal_likelihood}) is usually maximized to train the neural network. However, it is hard to directly optimize the log-likelihood function. Using an inference model $q(k)$ like VAE~\cite{kingma2013auto} to approximate the distribution of $\mathcal{K}$, especially $\sum_{k\in \mathcal{K}} q(k)=1$, we can re-write the log-likelihood function for one example as:
\begin{align}
    \log p(\mat{x}; \vct{\theta}) &=\sum_{k\in \mathcal{K}} q(k) \log p(\mat{x}; \vct{\theta}) \\
    &=\sum_{k\in \mathcal{K}} q(k)(\log p(\mat{x}, k; \vct{\theta})-\log p(k \vert \mat{x}; \vct{\theta})) \\
    &=\sum_{k\in \mathcal{K}} q(k) \log \frac{p(\mat{x}, k; \vct{\theta})}{q(k)}-\sum_{k\in \mathcal{K}} q(k) \log \frac{p(k \vert \mat{x}; \vct{\theta})}{q(k)} \\
    &=\operatorname{ELBO}(q, \mat{x} ; \vct{\theta})+\operatorname{KL}(q(k) \| p(k \vert \mat{x}; \vct{\theta})),
\end{align}
where $p(\mat{x}, k; \vct{\theta})=p(k \vert \mat{x} ; \vct{\theta}) p(\mat{x} ; \vct{\theta})$ so we have $\log p(\mat{x}; \vct{\theta})=\log p(\mat{x}, k; \vct{\theta})-\log p( k \vert \mat{x} ; \vct{\theta})$ and the evidence lower bound~(ELBO) is the lower bound of log-likelihood function since $\operatorname{KL}(q(k) \| p(k \vert \mat{x}; \vct{\theta}))\geq 0$.
When $\operatorname{KL}(q(k) \| p(k\vert\mat{x}; \vct{\theta}))=0$, the ELBO reaches its maximum value $\log p(\mat{x}; \vct{\theta})$, making $q(k)=p(k \vert \mat{x}; \vct{\theta})$. By replacing $q(k)$ with $p(k \vert \mat{x}; \vct{\theta})$ and ignoring the constant value $\sum_{k\in \mathcal{K}} -q(k)\log q(k)$, we are ready to maximize:
\begin{align}
    \sum_{k\in \mathcal{K}} p(k \vert \mat{x}; \vct{\theta}) \log p(\mat{x}, k; \vct{\theta}).\label{eq:lower_bound1}
\end{align}

\subsection{EM Framework}
\noindent\textbf{E-step.}\quad 
With the fixed $\vct{\theta}_t$ at the iteration $t$, this step aims to estimate $q_{t+1}(k)$ that makes $q_{t+1}(k)=p(k \vert \mat{x}; \vct{\theta}_t)$ so that $\operatorname{ELBO}(q_{t+1}, \mat{x} ; \vct{\theta}_t)=\log p(\mat{x}; \vct{\theta}_t)$.
Here, we perform the spherical $k$-means algorithm to estimate $p(k\vert \mat{x}; \vct{\theta}_t)$. We extract features from the target network since the target network performs more stable and yields more consistent clusters, similar to BYOL and MoCo. 

\noindent\textbf{M-step.}\quad 
With the fixed suboptimal $q_{t+1}(k)=p(k \vert \mat{x}; \vct{\theta}_t)$ after E-step, we turn to optimize the $\vct{\theta}$ to maximize the ELBO:
\begin{align}
\vct{\theta}_{t+1}=\argmax_{\vct{\theta}} \sum_{n=1}^{N} \operatorname{ELBO}(q_{t+1}, \mat{x}^{(n)}; \vct{\theta}). \label{eq:em_2}
\end{align}
Using a uniform prior for $k$ as $p(k)=1/K$, we can obtain $p(\mat{x}, k; \vct{\theta})=p(k)p(\mat{x}\vert k; \vct{\theta})= p(\mat{x}\vert k; \vct{\theta})/K$. By replacing $p(\mat{x}, k; \vct{\theta})$ in Eq.~(\ref{eq:lower_bound1}) and ignoring constant value, in this step, we should maximize:
\begin{align}
    \sum_{k\in \mathcal{K}} \mathbbm{1}(\mat{x} \in k) \log p(\mat{x}\vert k; \vct{\theta}), \label{eq:em_q}
\end{align}
where $p(k\vert \mat{x}; \vct{\theta})=\mathbbm{1}(\mat{x} \in k)$. $\mathbbm{1}(\cdot)$ is an indicator function using the hard labels estimated from E-step so that $\mathbbm{1}(\mat{x} \in k)=1$ if $\mat{x}$ belongs to $k$-th cluster; otherwise $\mathbbm{1}(\mat{x} \in k)=0$.
Following~\cite{li2020prototypical}, if we assume that the distribution for each cluster is the vMF distribution with a constant $\kappa$ as the temperature of softmax function, we can further obtain the follow:
\begin{equation}
    p(\mat{x}\vert k; \mat{\theta}) = \frac{\exp(\vct{\mu}_{k}\T \vct{v} / \tau)}{\sum_{k=1}^{K} \exp( \vct{\mu}_{k}\T \vct{v}/ \tau)}, \label{eq:px_ci}
\end{equation}
where $\vct{v}=f(\mat{x}; \vct{\theta})$, $\tau=1/\kappa$, and $\vct{\mu}_{k}$ is the cluster center of $k$-th cluster.
Combining Eqs.~(\ref{eq:em_q}) and~(\ref{eq:px_ci}), we can achieve the maximum log-likelihood estimation to find the optimal $\vct{\theta}^*$ by minimizing the following negative log-likelihood:
\begin{align}
\vct{\theta}^*=\argmin_{\vct{\theta}} \sum_{n=1}^{N} - \log \frac{\exp(\vct{\mu}_{y^{(n)}}\T \vct{v}^{(n)} / \tau)}{ \sum_{k=1}^{K} \exp(\vct{\mu}_{k}\T \vct{v}^{(n)} / \tau)}, \label{eq:elbo_3}
\end{align}
where $y^{(n)}$ is the pseudo-label for $\mat{x}^{(n)}$ estimated by the $k$-means algorithm in E-step.

Directly optimizing Eq.~(\ref{eq:elbo_3}) usually leads to improve the cluster compactness, which, however, will compromise the stability of BYOL since it does not consider the uniformity term. To this end, we propose a prototypical scattering loss~(\lossname) to maximize the log-likelihood at the cluster level by employing the clusters centers as the special instances, or prototypes from a set of instances. The \lossname is defined as:
\begin{align}
    \mathcal{L}_{\mathrm{psl}} = \frac{1}{K}\sum_{k=1}^{K} -\log \frac{\exp(\vct{\mu}_k\T \vct{\mu}_k^\prime / \tau)}{\exp(\vct{\mu}_k\T \vct{\mu}^\prime_k / \tau) + \sum_{j=1, j\neq k}^{K}\exp(\vct{\mu}_k\T \vct{\mu}_j / \tau)},
\end{align}
where $\{\vct{\mu}_1, \vct{\mu}_2, \ldots, \vct{\mu}_K\}$ and $\{\vct{\mu}_1^\prime, \vct{\mu}_2^\prime, \ldots, \vct{\mu}_K^\prime\}$ are $K$ prototypes from target and online networks, respectively. 
Here, instead of using the centroids computed from $k$-means, our cluster center $\vct{\mu}_k$ and $\vct{\mu}_k^\prime$ is empirically estimated within mini-batch $\mathcal{B}$ as follows:
\begin{align}
\vct{\mu}_{k} =\frac{\sum_{\mat{x} \in \mathcal{B} }  p(k\vert\mat{x}) f(\mat{x}) }{\|\sum_{\mat{x} \in \mathcal{B} }  p(k\vert\mat{x}) f(\mat{x})\|_{2}}, \quad\text{and}\quad
\vct{\mu}_{k}^\prime =\frac{\sum_{\mat{x} \in \mathcal{B} }  p(k\vert\mat{x}) f^\prime(\mat{x}) }{\|\sum_{\mat{x} \in \mathcal{B} }  p(k\vert \mat{x}) f^\prime(\mat{x})\|_{2}},\label{eq:cluster_create_appendix}
\end{align}
where $p(k\vert\mat{x})$ is estimated from E-step. When $K > \vert\mathcal{B}\vert$, it is obvious that the mini-batch cannot cover all clusters. To this end, we zero out the losses and logits of empty clusters for each iteration.

Intuitively, \lossname can encourage the prototypical alignment between two augmented views and the prototypical uniformity, hence maximizing the inter-cluster distance.

\subsection{Convergence Analysis} 
At E-step of the iteration $t$, we estimate $q_{t+1}(k)$ to make $\operatorname{ELBO}(q_{t+1}, \mat{x} ; \vct{\theta}_t)=\log p(\mat{x}; \vct{\theta}_t)$. At M-step after E-step, we obtain the optimized $\vct{\theta}_{t+1}$ with the fixed $q_{t+1}(k)$ so that $\operatorname{ELBO}(q_{t+1}, \mat{x}; \vct{\theta}_{t+1})\geq \operatorname{ELBO}(q_{t+1}, \mat{x}; \vct{\theta}_{t})$. Consequently, we obtain the following sequence:
\begin{align}
\log p(\boldsymbol{x} ; \vct{\theta}_{t+1}) \geq \operatorname{ELBO}(q_{t+1}, \boldsymbol{x} ; \vct{\theta}_{t+1}) \geq \operatorname{ELBO}(q_{t+1}, \boldsymbol{x} ; \vct{\theta}_{t})=\log p(\boldsymbol{x} ; \vct{\theta}_{t}).
\end{align}
Given $\log p(\boldsymbol{x} ; \vct{\theta}_{t+1}) \geq \log p(\boldsymbol{x} ; \vct{\theta}_{t})$, we can guarantee the convergence of our \methodname.
\clearpage
\section{Additional Experimental Results}
\label{sec:addition_results}

\begin{table*}[h]
  \centering
  \caption{
      Clustering results (\%) for fair comparisons. We train \methodname to demonstrate its effectiveness for fair comparisons with the following settings: (1) we exclude test set from the whole dataset; and (2) we use an original image size (224) for ImageNet-10 and ImageNet-Dogs. All results were trained with ResNet-34. There is no clear margin for CIFAR-10 and CIFAR-20 datasets with different splits, while significant improvements can be observed for ImageNet-10 and ImageNet-Dogs datasets. Considering that \methodname has already achieved state-of-the-art performance against previous work in Table 2, these results further demonstrate the superiority of \methodname.
  }
  \label{tab:fair_results_with_imagesize_and_split}
  
  \begin{tabular}{lllllll}
  \toprule
  & \multicolumn{1}{c}{NMI} & \multicolumn{1}{c}{ACC} & \multicolumn{1}{c}{ARI} & \multicolumn{1}{c}{NMI} & \multicolumn{1}{c}{ACC}      & \multicolumn{1}{c}{ARI} \\
  \midrule
  & \multicolumn{3}{c}{\textbf{CIFAR-10}}  & \multicolumn{3}{c}{\textbf{CIFAR-20}}                                                               \\
  \midrule
  Baseline (train+test)  & \multicolumn{1}{r}{\textbf{88.6}\std{1.0}} & \multicolumn{1}{r}{\textbf{94.3}\std{0.6}} & \multicolumn{1}{r}{\textbf{88.4}\std{1.1}} & \multicolumn{1}{r}{60.6\std{0.3}} & \multicolumn{1}{r}{61.4\std{1.1}} & \multicolumn{1}{r}{45.1\std{0.1}} \\
  Exclude test set & 88.3\std{0.2} & 94.2\std{0.2} & 88.1\std{0.3} & \textbf{61.2}\std{0.9} & \textbf{61.5}\std{0.9} & \textbf{45.9}\std{0.5} \\
  \midrule
  & \multicolumn{3}{c}{\textbf{ImageNet-10}} & \multicolumn{3}{c}{\textbf{ImageNet-Dogs}} \\ 
  \midrule
  Baseline~(96)  & 89.6\std{0.2} & 95.6\std{0.0} & 90.6\std{0.1} & 69.2\std{0.3} & 74.5\std{0.1} & 62.7\std{0.1} \\
  Large image size~(224) & \textbf{90.8}\std{0.4} & \textbf{96.2}\std{0.1} & \textbf{91.8}\std{0.3} & \textbf{73.7}\std{0.2} & \textbf{77.5}\std{0.1} & \textbf{67.5}\std{0.1} \\
  \bottomrule
  \end{tabular}
\end{table*}

\end{document}